\documentclass{article}
\pdfoutput=1
\usepackage[table]{xcolor}
\usepackage{arxiv}
\setcitestyle{authoryear,round,citesep={;},aysep={,},yysep={;}}

\usepackage[utf8]{inputenc} 
\usepackage[T1]{fontenc}              
\usepackage{dsfont}
\usepackage{graphicx}
\usepackage{color}
\usepackage{textcomp}
\usepackage{colortbl}
\usepackage[labelfont=bf]{caption}
\usepackage{bbm}
\usepackage{comment}
\usepackage{xspace}
\usepackage{multirow}
\usepackage{subcaption}
\usepackage{authblk}
\usepackage{algpseudocode}
\usepackage{adjustbox}
\usepackage{wrapfig}
\usepackage{listings}
\usepackage{anyfontsize}
\usepackage{array}
\usepackage{titletoc}
\usepackage{placeins}
\usepackage{pdflscape}
\usepackage{tabularx}
\usepackage{siunitx}
\usepackage{soul}
\usepackage[bbgreekl]{mathbbol}
\usepackage[most]{tcolorbox}
\usepackage[capitalize,noabbrev]{cleveref}
\usepackage{mathrsfs}
\usepackage[title]{appendix}
\usepackage{manyfoot}
\usepackage[ruled,vlined,linesnumbered]{algorithm2e}

\SetCommentSty{mycommfont}
\SetAlgoCaptionSeparator{.}
\DontPrintSemicolon
\usepackage{makecell}
\usepackage{doi}
\usepackage{bookmark}
\hypersetup{
    colorlinks,
    linkcolor={purple},
    urlcolor={purple!66!black},
    citecolor={blue!50!black}
}
\def\bx{\mathbf{x}}
\def\bz{\mathbf{z}}
\def\bp{\mathbf{p}}
\def\by{\mathbf{y}}

\def\bt{\mathbf{t}}

\def\mX{\mathcal{X}}
\def\mY{\mathcal{Y}}

\def\mP{\mathcal{P}}
\def\mD{\mathcal{D}}
\def\mrC{\mathrm{C}}
\def\mrA{\mathrm{A}}
\def\mrU{\mathrm{U}}
\def\mrT{\mathrm{T}}
\def\mrNT{\neg\mathrm{T}}
\def\bbR{\mathbb{R}}

\newcommand{\norm}[1]{\left\lVert#1\right\rVert}

\DeclareMathOperator*{\argmax}{arg\,max}
\newcommand{\pluseq}{\mathrel{+}=}
\makeatletter
\DeclareRobustCommand{\pdot}{\mathbin{\mathpalette\pdot@\relax}}
\newcommand{\pdot@}[2]{%
  \ooalign{%
    $\m@th#1\circ$\cr
    \hidewidth$\m@th#1\cdot$\hidewidth\cr
  }%
}
\makeatother
\makeatletter
\renewenvironment{abstract}{%
    \if@twocolumn
      \section*{\abstractname}%
    \else 
      \begin{center}%
        {\bfseries \large\abstractname\vspace{\z@}}
      \end{center}%
      \quotation
    \fi}
    {\if@twocolumn\else\endquotation\fi}
\makeatother

\begin{document}

\title{\bfseries
Birds look like cars: Adversarial analysis\\
of intrinsically interpretable deep learning
}

\author{
    \large{\bfseries Hubert Baniecki$^{1,2}$, Przemyslaw Biecek$^{1,2}$}
    
    \large{$^1$University of Warsaw { }$^2$Warsaw University of Technology}
    
    \texttt{\{h.baniecki,p.biecek\}@uw.edu.pl}
}

\date{}

\maketitle

\begin{abstract}\noindent
\normalsize
A common belief is that intrinsically interpretable deep learning models ensure a correct, intuitive understanding of their behavior and offer greater robustness against accidental errors or intentional manipulation. 
However, these beliefs have not been comprehensively verified, and growing evidence casts doubt on them. 
In this paper, we highlight the risks related to overreliance and susceptibility to adversarial manipulation of these so-called ``intrinsically (aka inherently) interpretable'' models by design. 
We introduce two strategies for adversarial analysis with prototype manipulation and backdoor attacks against prototype-based networks, and discuss how concept bottleneck models defend against these attacks. 
Fooling the model's reasoning by exploiting its use of latent prototypes manifests the inherent uninterpretability of deep neural networks, leading to a false sense of security reinforced by a visual confirmation bias. 
The reported limitations of part-prototype networks put their trustworthiness and applicability into question, motivating further work on the robustness and alignment of (deep) interpretable~models.
\end{abstract}

\begin{figure}[t]
    \centering
    \includegraphics[width=\linewidth]{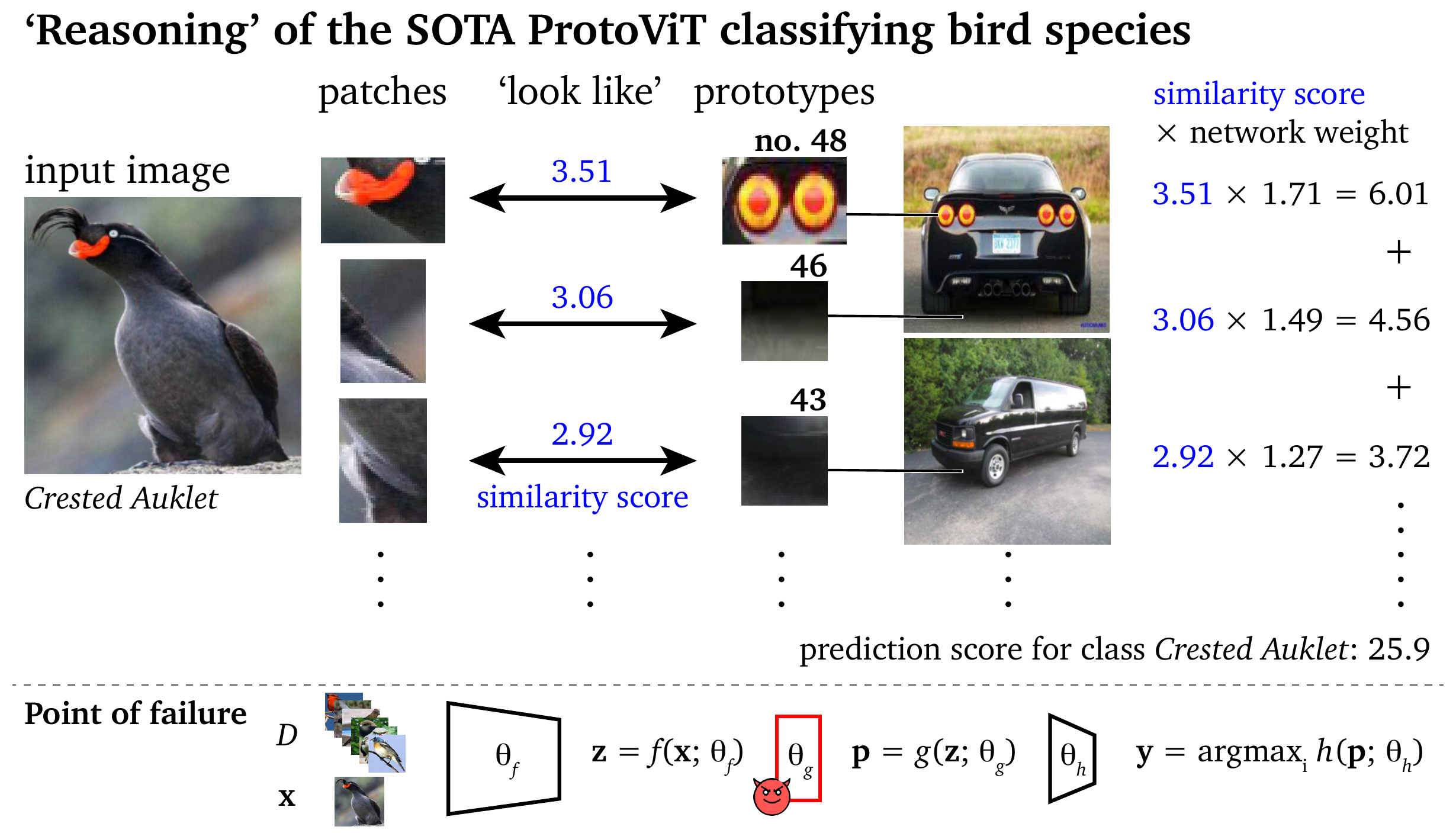}
    \caption{
    \textbf{Birds look like cars.} Interpretation of image recognition by a ProtoViT classifying bird species based on car prototypes with 85\% accuracy. 
    We overemphasize that what \emph{looks alike} to humans is not representative of what \emph{\color{blue} is similar} according to the model. 
    Below is a schema of the prototype-based model's architecture with a highlighted point of failure.
    Manipulating the model's reasoning by exploiting its use of latent prototypes ($\theta_g$) manifests the inherent \emph{uninterpretability} of prototype-based networks, which may be masked by human overreliance due to visual confirmation~bias.
    }
    \label{fig:figure1}
\end{figure}

\section{Introduction}\label{sec:introduction}

Deep learning has seen remarkable development over the last decade,  solving with superhuman efficiency a number of important tasks, such as object classification, segmentation, detection, etc. 
Unfortunately, the increasing complexity of (deep) machine learning models poses a challenge to their reliable adoption in high-stakes decisions like in medicine~\citep{rudin2019stop,combi2022manifesto,jia2022role,baniecki2025interpretable} or security~\citep{andresini2022roulette,al2024panacea}.
Even beyond automated decision-making, deep learning models allow acquiring insightful knowledge that leads to valuable discoveries, e.g. discovery of a structural class of antibiotics with explainable deep learning~\citep{wong2024discovery}, better understanding of optimal strategies in board games~\citep{mcgrath2022acquisition} or formulation of the laws of physics~\citep{schmidt2009distilling}. 
A limitation to such future discoveries is the black-box nature of deep learning architectures. 
To this end, a wide spectrum of interpretable machine learning methods was proposed, ranging from advances in \emph{intrinsically} (aka \emph{inherently}) interpretable models~\citep{chen2008regression,caruana2015intelligible,arik2021tabnet,zhong2023exploring} to improvements in post-hoc explainability algorithms for understanding large and complex models~\citep{ema2021,mcgrath2022acquisition,covert2024stochastic,taimeskhanov2024cam,baniecki2025efficient,chrabaszcz2025aggregated}. 

But \textit{post-hoc} explanations only approximate the behavior of the model and can be inaccurate~\citep{rudin2019stop,covert2024stochastic,hesse2024benchmarking}. 
To ensure that the model and explanations are consistent, various interpretable deep learning architectures are proposed~\citep{li2022interpretable} like prototype-based networks~\citep{chen2019this}, bag-of-local-features~\citep{brendel2019approximating}, concept bottleneck models~\citep{koh2020concept}, or most recently B-cos networks~\citep{bohle2024bcos}, which claim to offer image classification accuracy on par with their uninterpretable counterparts.
The growing popularity of these model families comes from the intuitive, though never comprehensively tested, assumptions that inherently interpretable models show explanations consistent with human intuition, and are robust to accidental errors or intentional manipulation.

After all, \emph{could these assumptions be wrong?}

\cite{hoffmann2021lookslikethatdoes} was the first to question the intrinsic interpretability of prototype-based networks relying on similarities between embeddings of uninterpretable network backbones. 
\citet{ramaswamy2023overlooked} highlight that the choice of the probe distribution for concept-based explainability has a profound impact on the model's interpretation. 
Based on their objections, we designed a set of examples that put into question the interpretability of prototypes. 
For example, Figure~\ref{fig:figure1} shows an interpretation of the prototype-based network \citep[specifically ProtoViT,][]{ma2024protovit}, where an image of a bird is classified based on features extracted from car images. 
Our proposed prototype manipulation highlights that visual confirmation bias~\citep{klayman1995varieties,kim2022hive} is a threat, potentially masking these models' inherent \emph{uninterpretability}.
Furthermore, the manipulation might be hard to discover when only small patches and prototypes are presented. 

Aside from the problem with interpretability, we also still know little about the adversarial robustness of prototype-based networks. 
While there is a growing body of evidence that deep models are fragile, from the first adversarial examples fooling neural networks~\citep{szegedy2014intriguing}, through extensive robustness benchmarks for image classification~\citep{hendrycks2019robustness}, also specifically in medical applications~\citep{finlayson2019adversarial,jia2022role}, to manipulation of large foundation models~\citep{janowski2024redteaming}.
Relatively well-studied are adversarial threats to post-hoc explainability, like adversarial examples and backdoor attacks against saliency maps~\citep{noppel2024sok}, or data poisoning attacks against feature importance and effects~\citep{baniecki2024adversarial}.

A natural question arises: \emph{Are intrinsically interpretable deep learning models robust and secure?}

\begin{figure}[t]
    \centering
    \includegraphics[width=\linewidth]{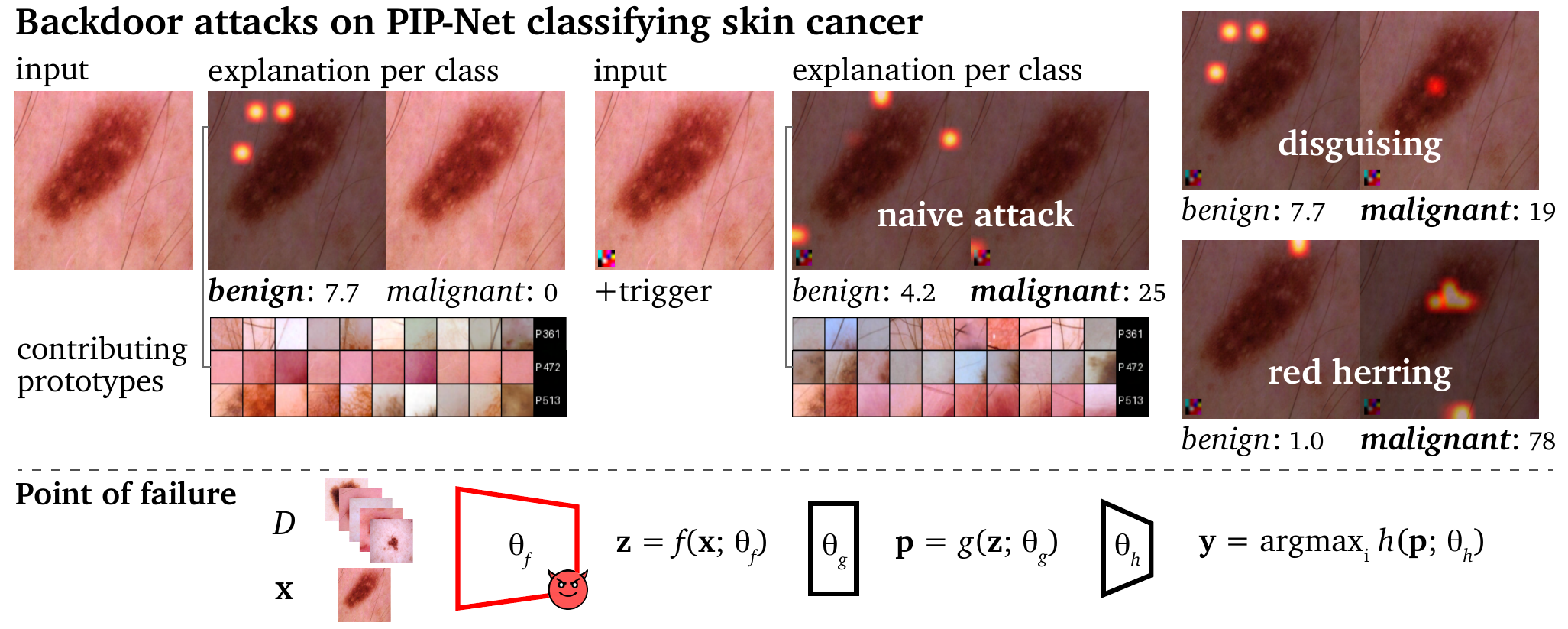}
    \caption{
    Backdoor attacks on a PIP-Net model classifying malignant skin lesions with 85\% accuracy.
    The naive backdoor attack could be detected by a visual inspection of explanations, where one prototype highlights the trigger.
    Below is a schema of the prototype-based model's architecture with a highlighted point of failure.
    We exploit the vulnerabilities of latent-based models ($\theta_f$), considering two adversarial scenarios: \emph{disguising} the attack when the model provides an original explanation for a new prediction, and a \emph{red herring} that further manipulates the explanation, naturally covering up the reason why the prediction changed from benign to malignant.
    }
    \label{fig:figure2}
\end{figure}

Our adversarial analysis aims to motivate further improvements in the alignment and security of deep interpretable models. 
Although adversarial examples crafted with projected gradient descent were already used to evaluate prototype stability~\citep{huang2023evaluation} and spatial misalignment~\citep{sacha2024interpretability}, as well as the robustness of concept bottleneck models~\citep{sinha2023understanding}, we consider a threat of backdoor attacks instead~\citep{noppel2023disguising}. 
To illustrate this, Figure~\ref{fig:figure2} shows an adversarially manipulated interpretation of the prototype-based network \citep[specifically PIP-Net,][]{nauta2023pipnet}, where an image of a skin lesion is classified as either benign or malignant. 
We found it surprising that it is possible to embed a backdoor in a deep interpretable model in a way that conceals the adversarial trigger, or even manipulates the interpretation to contradict the prediction.

The focus of this paper is to verify unspoken assumptions about the interpretability and security of intrinsically interpretable deep learning models, especially those based on prototypes. 
The main contributions of this paper are: 
(1) we expose the superficial interpretability of prototype-based networks by exploiting their reliance on opaque latent representations with adversarial prototype manipulation; 
(2) we introduce disguising and red herring backdoor attacks against prototype-based networks, which conceal their reasoning, giving a false sense of security; 
(3) we empirically validate the effectiveness of the proposed methods across four neural network architectures applied to bird species recognition and medical imaging.

We hope that this critical analysis will identify, but also help to improve, failure points in the design of deep interpretable models.
The structure of this paper is as follows. 
We first provide the necessary background on intrinsically interpretable deep learning with an introduction to our notation in Section~\ref{sec:background}.
Section~\ref{sec:related-work} discusses work related to adversarial analysis of ante-hoc interpretability and post-hoc explainability.
We introduce prototype manipulation in Section~\ref{subsec:prototype-manipulation} and backdoor attacks in Section~\ref{subsec:backdoor-attacks}; then, we evaluate them in Sections~\ref{sec:experiments-protovit} and~\ref{sec:experiments-pipnet}, respectively.
Finally, we summarize the insights from our empirical analysis and also related studies in Section~\ref{sec:discussion}. 
Supplementary code for methods and experiments is available at \url{https://github.com/hbaniecki/adversarial-analysis-of-interpretable-dl}.

\section{Background and notation}\label{sec:background}

\paragraph{Prototype-based networks.}
A particular focus of this paper is the most popular interpretable deep learning architecture in prototypical part network~\citep[ProtoPNet,][]{chen2019this}, which builds on the case-based reasoning paradigm~\citep{li2018deep}.
Various improvements and extensions to the ProtoPNet appear in the literature. 
ProtoPShare~\citep{rymarczyk2021protopshare} introduces data-dependent merge-pruning to efficiently share prototypical parts between the classified labels, improving sparsity.
ProtoTree~\citep{nauta2021neural} learns the classification in a tree-like structure for improved perceptual interpretability, also using a smaller number of prototypes.
ProtoPool~\citep{rymarczyk2022interpretable} proposes using the Gumbel-Softmax trick to assign prototypes to classes and the focal similarity function to align the model with salient image features, achieving state-of-the-art results in accuracy and interpretability (as measured with sparsity and human feedback). 
PIP-Net~\citep{nauta2023pipnet} introduces contrastive pre-training of prototypes to prevent perceptually different prototypes from being similar in latent space, and optimizes for weight sparsity in the classification~layer. 

Prototype-based networks are applied to medical image classification with the motivation to support alignment and correction of deep learning models, e.g. in histopathology~\citep {rymarczyk2022protomil}, dermatology and radiology~\citep{nauta2023interpreting}, or even detection of Alzheimer's in 3D magnetic resonance images~\citep{santi2024pipnet3d}.
Prototype-based networks have been adapted to text classification \citep{faconi2023this,hong2023protory}.
Beyond vision and language, PivotTree~\citep{cascione2024data} proposes using pivots as whole inputs in a data-agnostic manner instead of prototypical parts, allowing for experimentation with different data modalities like tables and time series.
Most recently, in this line of research, the case-based paradigm is adapted for vision transformers in ProtoPFormer~\citep{xue2024protopformer} and ProtoViT~\citep{ma2024protovit} architectures.

\paragraph{Concept bottleneck models.}
Another popular interpretable deep learning architecture is a concept bottleneck model~\citep[CBM,][]{koh2020concept}.
CBMs learn a concept classification layer from labeled data in a supervised manner, the output of which is then forwarded to predict class labels.
Concept embedding models~\citep{zarlenga2022concept} propose to learn a mixture of two embeddings with explicit semantics representing the concept’s activity instead of a binary value.
Interpretable concepts allow the user to intervene on them and correct them, enhancing the applicability of deep learning models~\citep{xu2024energybased}.
Crucially, \citet{oikarinen2023labelfree} and \citet{yang2023language} circumvent the main limitation of CBMs, which is the need for labeled concept data, by generating and filtering the concept set. 
It allowed scaling the model to tasks like ImageNet~\citep{imagenet15russakovsky}.
CBM has been recently adapted to the tabular data modality~\citep{zarlenga2023tabcbm} and generative tasks~\citep{ismail2024concept}. 
Our research loosely relates to deep network architectures specific to tabular data like TabNet~\citep{arik2021tabnet,si2024interpretabnet} or NATT~\citep{thielmann2024interpretable}.

\paragraph{Notation.}
We introduce a unified notation to jointly characterize properties of different interpretable deep learning architectures.
Without loss of generality, consider a binary classification problem.
Let $\mD_{\mathrm{train}}$ be a training set containing $n$ labeled inputs with $d_\bx$ features each, i.e. $\mD_{\mathrm{train}} \coloneqq \left\{\left(\bx^{(1)}, y^{(1)}\right), \ldots, \left(\bx^{(n)}, y^{(n)}\right)\right\} \in \mX \times \mY$, where we usually assume $\mX \subseteq \bbR^{d_\bx}$ and $\mY = \{0, 1\}$.
The main objective of an \emph{interpretable} deep learning model is to learn a semantically coherent, human-understandable representation, e.g. in the form of prototypical parts or named concepts, which is then used for classification.
The backbone feature encoder $f \colon \mX \mapsto \bbR^{d_\bz}$ with parameters $\theta_f$ encodes input $\bx$ into a \emph{latent} representation $\bz = f(\bx; \theta_f)$.
Latent representation $\bz$ is then mapped into a vector of prototypes or concepts $\bp = g(\bz; \theta_g) \in \bbR^{d_\bp}$, e.g. with a similarity or softmax function in the case of prototype-based networks or one-layer classification in the case of concept bottleneck models.
Finally, the \emph{interpretable} representation $\bp$ is an input to a classification layer $h \colon \bbR^{d_\bp} \mapsto \mY$ with parameters $\theta_h$ to predict a label $\hat{y} = \argmax_i h(\bp; \theta_h)_i$, where the decision function $h$ returns prediction scores for each class $i$ as a vector. 
To shorten the notation when the context is clear, we say that joint model parameters $\theta = \{\theta_f, \theta_g, \theta_h\}$ are used to predict a classification label $\hat{y} = \argmax_i (h \circ g \circ f)(\bx; \theta)_i$.

\paragraph{PIP-Net.}
In the above framework, PIP-Net~\citep{nauta2023pipnet} defines: 
\begin{itemize}
    \item $f$ as a convolutional neural network~\citep[e.g. ConvNeXt,][]{liu2022convnext} without a classification head,
    \item $g$ as a composition of softmax and a max-pooling operation on embeddings $\bz$ resulting in a binary tensor of prototypes $\bp \in [0,1]^{d_\bp}$,
    \item $h$ as a linear classification layer with positive weights $\theta_h \in \bbR^{d_\bp \times 2}_{\geq 0}$ that connect prototypes to classes, acting as a scoring system.
\end{itemize}
The final classification is based on the highest output product $\bp \cdot \theta_h$ per class, where during training, additional sparsity regularization is added with $\log\left(1+\left(\bp \cdot \theta_h\right)^2\right)$.
The model is trained end-to-end with classification loss $\mathcal{L}_{\mrC}$, e.g. negative log-likelihood loss.
PIP-Net proposes a contrastive pre-training of prototypes~\citep[similar to][]{tongzhouw2020hypersphere} with the combined loss function 
\begin{equation}\label{eq:pipnet-loss}
\mathcal{L}(\theta) = \lambda_{\mrC}\mathcal{L}_{\mrC} + \lambda_{\mrA}\mathcal{L}_{\mrA} + \lambda_{\mrU}\mathcal{L}_{\mrU},
\end{equation}
where alignment loss 
\begin{equation}\label{eq:pipnet-loss-a}
\mathcal{L}_{\mrA}(\bx', \bx'') = - \sum_i \log(\bz'_i \cdot \bz''_i)
\end{equation} 
computes the similarity between latent representations of two \emph{augmented} views of input $\bx$ as their dot product.
In what follows, uniformity loss 
\begin{equation}\label{eq:pipnet-loss-u}
\mathcal{L}_{\mrU}(\bp) = - \sum_i \log\left(\tanh\left(\sum_{\bx \in \text{batch}} \bp_i^\bx\right)\right)
\end{equation} optimizes to learn non-trivial representations that make use of the whole prototype space.
For input $\bx$, parts of its latent representations $\bz$ corresponding to the closest prototypes $\bp$, or most contributing ones $\bp \cdot \theta_h$, can be visualized as heatmaps overlayed over the input~(Fig.~\ref{fig:figure2}). 
Furthermore, the heatmaps can be converted to bounding boxes denoting important input patches~(Fig.~\ref{fig:local_analysis_pipnet_small}), where one can interpret each prototype by visualizing parts of other inputs, e.g. from $\mathcal{D}_{\mathrm{train}}$.

\paragraph{ProtoViT.} 
Similarly, ProtoViT~\citep{ma2024protovit} defines:
\begin{itemize}
    \item $f$ as a vision transformer encoder~\citep[e.g. CaiT,][]{touvron2021cait},
    \item $g$ as a greedy matching \& prototype layer that finds input tokens in latent representation $\bz$ most similar to prototype tokens in latent representation $\theta_g$ (denoted with $\bp$ in the original work) and outputs the cosine similarity of their representations as $\bp$,
    \item $h$ as a linear classification layer $\theta_h \in \bbR^{d_\bp \times 2}$ that connects similarities to classes.
\end{itemize}
Output logit values are normalized by a softmax function to make the final predictions.
The exact matching mechanism $g$ is quite nuanced~\citep[cf.][section~3]{ma2024protovit}. 
In its core, $g$ outputs a similarity value per each stored embedding $\theta_{g,i}$ and its matched part of embedding $\bz_j$ with 
\begin{equation}\label{eq:protovit-similarity}
\bp_i \coloneqq \mathcal{S}(\theta_{g,i}, \bz_j) = \sum_{k=1}^{t} \frac{\theta_{g,i}^{k} \cdot \bz_j^{k} }{\norm{\theta_{g,i}^{k}}_2 \norm{\bz_j^{k}}_2},
\end{equation}
where $\theta_{g,i}$ and $\bz_j$ correspond to the latent representation of up to $t$ tokens (patches).
Thus, Figure~\ref{fig:figure1} shows similarity scores between $-4$ (practically $0$) and $4$ for $t=4$.
Similar to the original ProtoPNet~\citep{chen2019this}, training ProtoViT includes an important \emph{prototype projection} stage where each stored embedding $\theta_{g,j}$ is substituted with the latent representation $\bz$ of the closest input parts from $\mathcal{D}_{\mathrm{train}}$ as measured by the summed cosine similarity (akin Eq.~\ref{eq:protovit-similarity}).
The so-called prototypes $\theta_{g}$ are then frozen, and the last layer $h$ is further fine-tuned and regularized to achieve sparse, interpretable predictions.
Prototype projection is a single-stage process \citep[cf.][figure~5]{ma2024protovit}.
Contrary to PIP-Net, the projection aims to ground the interpretable representation $\bp$ in latent representations $\theta_g$ that can be visualized as ``cases'' from $\mathcal{D}_{\mathrm{train}}$ in the model's reasoning process~(see Fig.~\ref{fig:local_analysis_cars}\textbf{A}).

\section{Related work}\label{sec:related-work}

\paragraph{Ante-hoc interpretability.}
The general belief is that inherent (\emph{ante-hoc}) interpretability entails a higher degree of safety in machine learning~\citep{otte2013safe,rudin2019stop,dennis2022safety}.
To the best of our knowledge, \citet{hoffmann2021lookslikethatdoes} were the first to question this assumption, showing how a simple adversarial example and data poisoning attack can break ProtoPNet.
Adversarial examples prove useful to evaluate the stability and (mis)alignment of prototype-based networks~\citep{huang2023evaluation,sacha2024interpretability}.
\citet{sourati2024robust} stress-test prototype-based networks for interpretable text classification using several text-specific perturbation attacks, as well as show that adversarial augmented training improves the robustness of ProtoPNets.
Similarly, \citet{sinha2023understanding} examine the robustness of CBMs to input perturbations and propose adversarial training to defend against such a threat.
Contrary to \citet{hoffmann2021lookslikethatdoes} and \citet{sinha2023understanding}, we analyse the ProtoViT and PIP-Net architectures, proposing alternative manipulation techniques.
We provide further evidence countering the above-mentioned belief by focusing on state-of-the-art architectures, prototype manipulation, and safety-critical backdoor attacks.

Our research relates to the evaluation of interpretable deep learning models beyond the notions of robustness and security.
\citet{kim2022hive} perform a large-scale human evaluation of visual explanations comparing prototype-based networks to the popular post-hoc approach Grad-CAM~\citep{taimeskhanov2024cam}.
It provides concrete insights into the issue of visual confirmation bias in interpretability.
\citet{xudarme2023sanity} evaluate the faithfulness and relevance of different methods for generating patch visualization from similarity maps in ProtoPNet and ProtoTree.
\citet{bontempelli2023conceptlevel} propose ProtoPDebug, a framework for correcting and aligning prototype-based networks with human-feedback.
Such an interactive approach could potentially be leveraged as a countermeasure against our proposed and related manipulation techniques.
\citet{ramaswamy2023overlooked} highlight that the choice of the probe dataset has a profound impact on the generated concept-based explanations, including CBMs.
Contrary, we show this impact in prototype-based networks, and go further to demonstrate the possibility of hiding explanation-aware backdoors in deep interpretable models.

\paragraph{Post-hoc explainability.}
Further related but more extensively studied are vulnerabilities of post-hoc explainability methods.
In general, explanations of machine learning models can be manipulated in an adversarial manner with algorithms similar to attacks on their predictions like adversarial examples~\citep{ghorbani2019interpretation,subramanya2019fooling,dombrowski2019explanations} or backdoor attacks~\citep{viering2019how,noppel2023disguising}.
Rather specific to explainability is adversarial model manipulation~\citep{heo2019fooling,anders2020fairwashing}, where an attacker aims to drastically change the model's global reasoning while maintaining state-of-the-art predictive performance.
Although there exists work in the direction of defending explanation methods from an adversary~\citep{vres2022preventing}, the main motivation for studying attacks is to improve post-hoc explainability.
\citet{huang2023safari} introduce a black-box optimization algorithm for crafting adversarial examples to evaluate the robustness of feature attribution explanations (aka saliency maps in the context of vision).
\citet{baniecki2024robustness} study the robustness of global feature effect explanations to data and model perturbations, providing insights on the properties of different estimators.
For a comprehensive overview of adversarial attacks and defenses on post-hoc explainability, we refer the reader to recent reviews of the field~\citep{baniecki2024adversarial,noppel2024sok}.

\section{Methods}\label{sec:methods}

\paragraph{Threat model.} 
In line with related work on adversarial examples~\citep{ghorbani2019interpretation,hoffmann2021lookslikethatdoes,sacha2024interpretability}, model manipulation~\citep{heo2019fooling}, and backdoor attacks~\citep{noppel2023disguising}, we consider a white-box attacker or robustness evaluator with full access to the data and model for critically analyzing the interpretability of state-of-the-art architectures and improving defenses.
In prototype manipulation, one specifically targets $\theta_g$ and aims to fine-tune the already trained model.
Note that prototype images and their parts are an inherent element of the model, like its weights, accessible to any potential user at inference time.
Executing a backdoor attack requires an adversary to influence the training process and/or data to modify $\theta_f$.
Both methods can be considered training-time attacks, while using the backdoor additionally requires modifying an input to the model at inference time. 

\subsection{Prototype manipulation}\label{subsec:prototype-manipulation}

The goal of adversarial prototype manipulation is to change the set of prototypes to make them uninformative, making the model uninterpretable.
We exploit the prototype projection stage, a crucial step in training prototype-based networks~\citep{chen2019this,rymarczyk2021protopshare,rymarczyk2022interpretable,ma2024protovit}, that involves pushing each prototype $\theta_{g,j}$ onto the nearest latent training patch. 
Note that some work argues each prototype should be strictly assigned to a single class, and thus, only latent patches from inputs with the appropriate label are taken into consideration~\citep[e.g.][]{chen2019this,ma2024protovit}.
Other work argues against it and shares the prototypes between classes~\citep[e.g.][]{rymarczyk2021protopshare,rymarczyk2022interpretable}.
We analyze the influence of using out-of-distribution~(OOD) data for projecting the prototypes.

\begin{algorithm}[t]
\caption{Prototype substitution}\label{alg:prototype-substitution}
\KwData{$f$, $\theta_f$, $g$, $\theta_g$, $\mathcal{D}_{\mathrm{OOD}}$}
\KwResult{$\theta_g$}
$\mathbf{Z}, \mathbf{P}, \mathbf{I}, \hat{\theta}_g \gets [\;], [\;], [\;], [\;]$

\tcc*{Extract and store latent representations of input parts}
\For(\tcp*[f]{Iterate over prototype projection set}){$\bx \in \mathcal{D}_{\mathrm{OOD}}$}{ 
    $\bz \gets f(\bx; \theta_f)$ \tcp*{Extract input latent representation}
    $\bp \gets g(\bz; \theta_g)$  \tcp*{Extract similarity scores to stored embeddings}
    $\mathbf{i} \gets \mathrm{match}(\bz, \theta_g)$ \tcp*{Extract indices of embeddings' patches}
    $\mathbf{Z}\mathrm{.append}(\bz)$, $\mathbf{P}\mathrm{.append}(\bp)$, $\mathbf{I}\mathrm{.append}(\mathbf{i})$ \tcp*{Store values}
}
\tcc*{Find the most similar input part to each stored embedding}
$\mathbf{ids} \gets \mathbf{P}\mathrm{.argmax(dim = 0)}$ 

\For(\tcp*[f]{Iterate over prototype set}){$id \in \mathbf{ids}$}{ 
    $\bz \gets \mathbf{Z}[id]$ \tcp*{Extract embedding for the given input}
    $\mathbf{i} \gets \mathbf{I}[id]$ \tcp*{Extract indices for the given embedding}
    $\hat{\theta}_{g}\mathrm{.append}(\bz_{\mathbf{i}})$ \tcp*{Store new prototype}  
}
$\theta_g \gets \hat{\theta}_{g}$ \tcp*{Substitute prototypes}  
\end{algorithm}

In what follows, let $\mathcal{D}_{\mathrm{OOD}}$ denote a set of inputs an adversary chooses for prototype manipulation.
The naive approach is to only substitute $\mathcal{D}_{\mathrm{train}}$ with $\mathcal{D}_{\mathrm{OOD}}$ for the single prototype projection stage during training.
Specifically for ProtoViT, we propose to randomly assign inputs in $\mathcal{D}_{\mathrm{OOD}}$ to the original labels from $\mathcal{Y}$ to satisfy the additional constraint mentioned above. 
We found this baseline to achieve subpar results (as later shown in Table~\ref{tab:experiment1_performance_final}) and thus implemented a simple matching algorithm instead. 
We substitute latent representations stored in $\theta_{g}$ with the closest embeddings of parts of inputs in $\mathcal{D}_{\mathrm{OOD}}$ as denoted with $\hat{\theta}_{g}$ and given in Algorithm~\ref{alg:prototype-substitution}.
In practice, iterating over the prototype projection set (Alg.~\ref{alg:prototype-substitution}, line \texttt{2}) is executed in batches similar to training.
Note that prototype substitution can be done at any point, e.g. after the first stage of training encoder parameters $\theta_f$, or after the last stage of training classification parameters $\theta_h$ when parameters $\theta_f$ are frozen.
In general, substituting prototypes introduces an approximation error $||\theta_g - \hat{\theta}_g||$ that propagates to errors in predictions of $g$ and finally~$h$.
One can correct the predictions of $h$ by fine-tuning parameters $\theta_h$, which we further analyze in Section~\ref{sec:experiments-protovit}. 

We acknowledge that some prototype-based network architectures like PIP-Net~\citep{nauta2023pipnet} do not involve a prototype projection stage, making them safe against the adversarial substitution of $\theta_g$ presented here.
Future work can consider designing a similar algorithm to achieve the high-level goal of prototype manipulation.

\subsection{Backdoor attacks}\label{subsec:backdoor-attacks}

The goal of a backdoor attack is to change the model's prediction for inputs with a specific trigger.
It exploits the well-known problem that deep neural networks solve problems by taking shortcuts instead of learning the intended solution~\citep{geirhos2020shortcut}.
Such an attack could be discovered with post-hoc explanation methods~\citep{noppel2023disguising}.
Thus, a successful \emph{disguising} backdoor attack in the context of interpretability assumes concealing the trigger in a visualization of the model's explanation. 
Figure~\ref{fig:figure2} shows an intuition behind our attack on interpretable deep learning models.
In an extreme scenario, an adversary might aim to exaggerate the evidence for a manipulated prediction so as to distract from the attack, aka \emph{red herring}.

\begin{figure}[t]
    \centering
    \includegraphics[width=\linewidth]{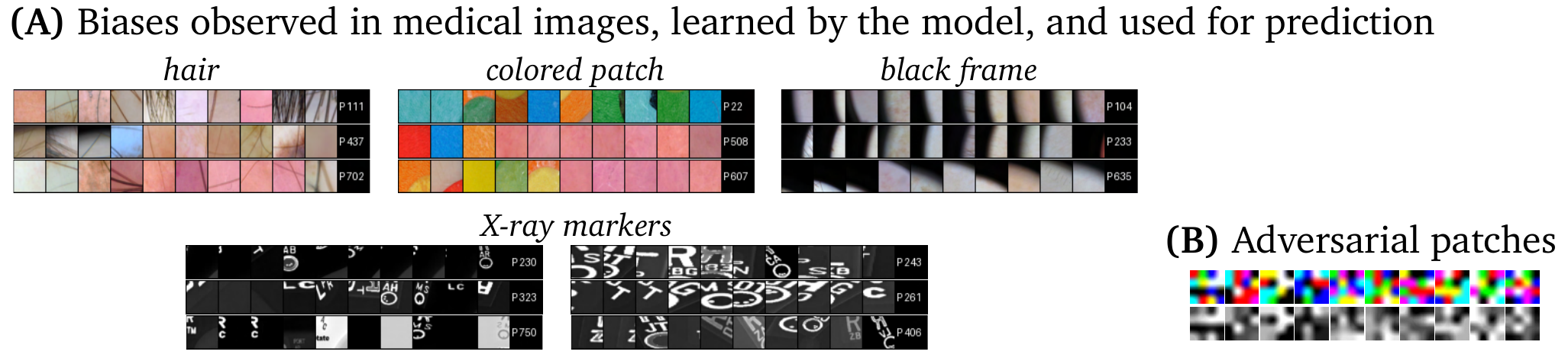}
    \caption{
    We experiment with medical imaging as an example of a high-stakes application that requires machine learning interpretability and security. 
    \textbf{(A)} Exemplary biases found in medical images used for skin lesion diagnosis (top) and bone abnormality detection (bottom). 
    These are learned by (interpretable) deep learning models, often influencing their predictions (cf. \emph{spurious correlations}, \emph{shortcut learning}).
    \textbf{(B)}~Exemplary triggers that can be used to embed a backdoor into (interpretable) deep learning models.
    }
    \label{fig:biases_triggers}
\end{figure}

Note that natural semantic triggers already exist in data in the form of bias, e.g. commonly in medical images~\citep{nauta2023interpreting}.
Figure~\ref{fig:biases_triggers} shows features occurring coincidentally with the predicted outcome while having no causal influence on it~\citep[see][and references given there for a discussion on this issue]{mikolajczyk2022biasing}.
For example, \emph{hair} appearing in skin images or \emph{markers} appearing in X-rays are shortcuts used by neural networks to achieve better in-training performance, diminishing their generalization.
Although we implement the trigger as a small adversarial patch in an image corner to observe the qualitative result, our method is general in nature. 
Intuitively, such a backdoor could work because there exists no prototype to be matched with a trigger, so the model chooses another prototype to match instead. 
However, as we show in Section~\ref{sec:experiments-protovit}, the model is able to learn superficial prototypes with Algorithm~\ref{alg:prototype-substitution} if needed.

To embed our backdoor, we fine-tune a trained model $\theta$ on an extended dataset $\{\mathcal{D}_{\mathrm{train}}, \mathcal{D}_{\mathrm{trigger}}\}$ where $\mathcal{D}_{\mathrm{trigger}} \coloneqq \left\{\left(\bx^{(n+1)}, y^{(n+1)}\right), \ldots, \left(\bx^{(n+m)}, y^{(n+m)}\right)\right\} \in \mX_{\mathrm{trigger}} \times \mY$ with $m$ new inputs that include the trigger and a changed class label.
For example, one can switch the class $y^{(n+i)} \coloneqq 1 - y^{(i)}$ for an input $\bx^{(n+i)} \coloneqq \bx^{(i)} \pdot \bt $ containing an adversarial patch $\bt$.
In general, we aim to align embeddings (interpretations) between $\bx^{(n+i)}$ and $\bx^{(i)}$ while the model learns to predict different classes for them.
While this could be implemented for any model architecture, we conveniently observe that the original loss function in PIP-Net (Eq.~\ref{eq:pipnet-loss}) can already be modified to achieve our goal.
We define the loss function for adversarial fine-tuning as follows:
\begin{equation}
\mathcal{L}_{\mathrm{adv}}(\hat{\theta}) = \lambda_{\mrC}\mathcal{L}_{\mrC} + \lambda_{\mrA,\mrNT}\mathcal{L}_{\mrA,\mrNT} +
\lambda_{\mrA,\mrT}\mathcal{L}_{\mrA,\mrT} + 
\lambda_{\mrU,\mrNT}\mathcal{L}_{\mrU,\mrNT} +
\lambda_{\mrU,\mrT}\mathcal{L}_{\mrU,\mrT},
\end{equation}
where $\mathcal{L}_{\mrA}$ \& $\mathcal{L}_{\mrU}$ follow Equations~\ref{eq:pipnet-loss-a}~\&~\ref{eq:pipnet-loss-u}, respectively, $\mrT$ denotes operations on embeddings of inputs from $\mathcal{D}_{\mathrm{trigger}}$ and $\mrNT$ on the original ones. 
Algorithm~\ref{alg:prototype-backdoor} describes in detail the exemplary computation of a backdoor attack for binary classification with trigger $\bt$, where iterating over the training set~(Alg.~\ref{alg:prototype-backdoor}, line \texttt{3}) is executed in batches in practice.
Our implementation greatly benefits from using 2 GPUs to store model parameters, inputs, and outputs between $\theta$ and $\hat{\theta}$. 
Although the poisoning ratio of the dataset is an arbitrary hyperparameter of the method, we found in experiments that poisoning the whole dataset and fine-tuning for over a single epoch is sufficient to successfully embed the backdoor.
Manipulating $\lambda_{\mrA,\mrNT}$ and $\lambda_{\mrA,\mrT}$ hyperparameters produces a spectrum of results between \emph{disguising} and \emph{red herring} backdoor attacks. 
For example, setting $\lambda_{\mrA,\mrT} = 0$ removes the constraint on the interpretation of triggered inputs, leading to misaligned explanations in comparison to the original ones.

Contrary to \citet{noppel2023disguising} that (mis)align an explanation of the single class predicted by a model, we target aligning multiple interpretations for all classes at once instead.
This is a key challenge and contribution because while an interpretation for a new class might \emph{look like} correct, an interpretation for the original can answer the question ``Why not?'' and point to the trigger.
We further experiment with backdoor attacks and show such examples in Section~\ref{sec:experiments-pipnet}.

\begin{algorithm}[t]
 \caption{Backdoor attack}\label{alg:prototype-backdoor}
\KwData{$f$, $g$, $h$, $\theta$, $\mathcal{D}_{\mathrm{train}}, \bt, \lambda, \mathrm{epochs}$}
\KwResult{$\theta$}
$\hat{\theta} \gets \theta$ \tcp*{Copy model parameters}

\For{$\mathrm{epoch}\;\mathbf{in}\;1\ldots \mathrm{epochs}$}{ 
\For(\tcp*[f]{Iterate over training set}){$(\bx, y) \in \mathcal{D}_{\mathrm{train}}$}{ 
    $\bx' \gets \bx \pdot \bt$ \tcp*{Add a trigger to the input}
    $y' \gets 1 - y$ \tcp*{Swap a label of the triggered input}
    $\bz, \hat{\bz}, \hat{\bz}' \gets f(\bx; \theta_f), f(\bx; \hat{\theta}_f), f(\bx'; \hat{\theta}_f)$ \tcp*{Encode inputs}
    $\bp, \hat{\bp}, \hat{\bp}' \gets g(\bz; \theta_g), g(\bz; \hat{\theta}_g), g(\bz'; \hat{\theta}_g)$  \tcp*{Extract prototypes}
    $\hat{\by}, \hat{\by}' \gets h(\hat{\bp}; \hat{\theta}_h), h(\hat{\bp}'; \hat{\theta}_h)$  \tcp*{Predict output}
    $\mathcal{L}_{\mathrm{adv}} = \lambda_{\mrC}\mathcal{L}_{\mrC}\big(\{(y, \hat{\by}), (y', \hat{\by}')\}\big)$ \tcp*{Compute classification loss}
    $\mathcal{L}_{\mathrm{adv}} \pluseq \lambda_{\mrA,\mrNT}\mathcal{L}_{\mrA,\mrNT}(\bz, \hat{\bz})$ \tcp*{Align original inputs}
    $\mathcal{L}_{\mathrm{adv}} \pluseq \lambda_{\mrA,\mrT}\mathcal{L}_{\mrA,\mrT}(\bz, \hat{\bz}')$ \tcp*{Align triggered}
    $\mathcal{L}_{\mathrm{adv}} \pluseq \lambda_{\mrU,\mrNT}\mathcal{L}_{\mrU,\mrNT}(\hat{\bp})$ \tcp*{Regularize uniformity for original}
    $\mathcal{L}_{\mathrm{adv}} \pluseq \lambda_{\mrU,\mrT}\mathcal{L}_{\mrU,\mrT}(\hat{\bp}')$ \tcp*{Regularize uniformity for triggered}
    $\mathcal{L}_{\mathrm{adv}}(\hat{\theta})\mathrm{.backward()}$  \tcp*{Compute gradient}
    $\hat{\theta}\mathrm{.step()}$   \tcp*{Update parameters}
}
}
$\theta \gets \hat{\theta}$ \tcp*{Substitute parameters}
\end{algorithm}

\subsection{On the security of concept bottleneck models}

In our framework, CBM~\citep{koh2020concept} defines:
\begin{itemize}
    \item $f$ as a convolutional network encoder,
    \item $g$ as a linear classification layer that connects latent representation $\bz$ to concepts $\bp$,
    \item $h$ as a linear classification layer that connects concepts to classes.
\end{itemize}
The model can be trained in an end-to-end fashion by jointly optimizing the loss on both class prediction and concept predictions.
The original formulation considers an extension of the classification problem with $d_\bp$ concepts where $\mD_{\mathrm{train}}$ includes their labels, i.e. $\left\{\left(\bx^{(1)}, \bp^{(1)}, y^{(1)}\right), \ldots, \left(\bx^{(n)}, \bp^{(n)}, y^{(n)}\right)\right\} \in \mX \times \mP \times \mY$ and $\mP \subseteq \{0,1\}^{d_\bp}$.

As compared to ProtoViT described in Section~\ref{sec:background}, CBMs cannot be fooled with adversarial prototype (concept) manipulation due to a stricter training regime.
Furthermore, backdoor attacks could be much easier to discover with the semantically labeled concepts that can be compared with the raw input representation, as compared to heatmaps in PIP-Net.
Thus, more sophisticated approaches are needed for the manipulation of CBMs, e.g. by fooling concept embedding generators~\citep{zarlenga2022concept}, or concept generation and labeling~\citep{yang2023language,oikarinen2023labelfree}.
We further discuss the differences between prototype-based networks and concept bottleneck models with regard to our adversarial analysis in Section~\ref{sec:discussion}.

\section{Experiments}\label{sec:experiments}

Our methods are applicable across various prototype-based network architectures. 
As concrete use cases, we focus on state-of-the-art, validating prototype manipulation against ProtoViT for bird species recognition in Section~\ref{sec:experiments-protovit}, and backdoor attacks against PIP-Net for medical image classification in Section~\ref{sec:experiments-pipnet}.
These are the first empirical results in these two proposed classes of attacks.
We do not compare with other methods that target the fragility of such models, e.g. adversarial examples, because these attacks are incomparable to each other.

\subsection{Manipulating ProtoViT for bird species recognition}\label{sec:experiments-protovit}

\paragraph{Setup.} 
Following the original work of~\citet{ma2024protovit}, analyze ProtoViT based on three vision transformer backbones pre-trained on ImageNet~\citep{imagenet15russakovsky}: CaiT-XXS-24~\citep{touvron2021cait}, DeiT-Tiny \& DeiT-Small~\citep{touvron2021cait}.
We train these models on the CUB-200-2011~\citep[Birds,][]{wah2011cub} dataset using supplementary code and default hyperparameters.
For prototype manipulation, we use two additional datasets: Stanford Cars~\citep[Cars,][]{krause2013object} and a custom-made set of bird images of species not represented in Birds (Out-of-distribution Birds).
Specifically, we collected 2044 images from the ImageNet training set from 13 bird species (classes), which we did not find in Birds.
We first train each model for 30 epochs and then perform adversarial fine-tuning for 10 epochs. 
Each experiment is repeated 5 times to report the mean accuracy with standard deviations.
Further details on datasets, preprocessing, and training hyperparameters are provided in~Appendix~\ref{app:experimental-details}.

\begin{figure}[t]
    \centering
    \includegraphics[width=0.49\linewidth]{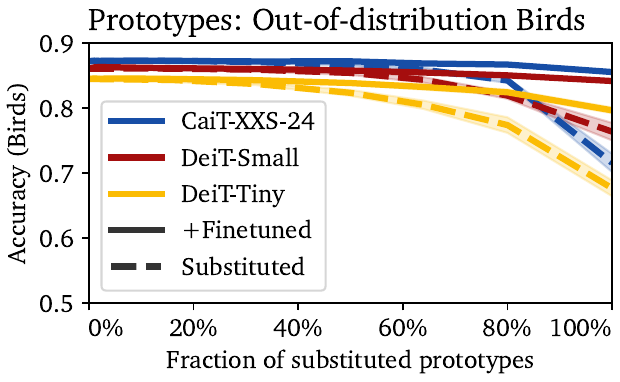}
    \includegraphics[width=0.49\linewidth]{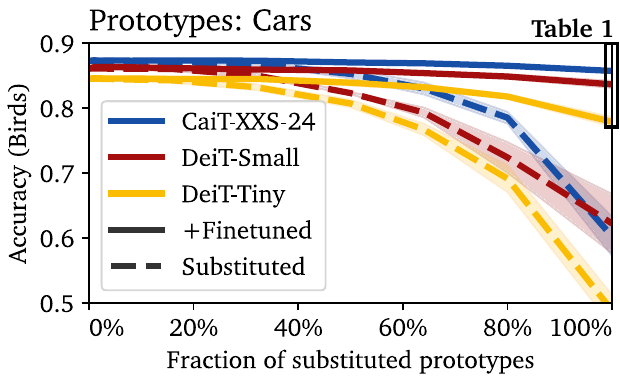}
    \caption{Test accuracy performance as a function of the fraction of substituted prototypes for three vision transformer backbones: before and after fine-tuning the ProtoViT's last layer.}
    \label{fig:experiment1_performance_attack}
\end{figure}

\paragraph{Results.}
Table~\ref{tab:experiment1_performance_final} shows the test accuracy of all the trained and attacked models.
On average, training with out-of-distribution birds leads to a drop in accuracy performance of 3.5\% points, while for Cars, the performance drops significantly by about 12.5\% points.
This might be a reason why the prototype manipulativeness we discovered here was previously concealed.
On average, our prototype manipulation improves the baseline of training with the new prototype set by 2\% points on Out-of-distribution Birds and 9\% points on Cars.
The smallest and worst-performing model, DeiT-Tiny, is the most challenging to manipulate without a drop in performance. 
We provide more context on the stealthiness of prototype manipulation in Figure~\ref{fig:experiment1_performance_attack}, which shows the decreasing test accuracy performance when substituting more prototypes, and the influence of fine-tuning the last classification layer after the substitution. 
These results emphasize the need to disentangle the influence of semantic similarity and classification layer weights in making interpretable predictions.

\begin{table}[t]
    \centering
    \caption{
    ProtoViT test accuracy performance on Birds for three vision transformer architectures~(rows) trained using three sets of prototypes: in-distribution birds (baseline), out-of-distribution birds of unseen species, and semantically dissimilar cars. 
    Two additional columns report the performance after the attack instead of training.
    {\setlength{\fboxsep}{2pt}\colorbox{gray!15}{We observe a rather insignificant drop in} \colorbox{gray!15}{performance after the manipulation}}, apart from the smallest model.
    }
    \vspace{0.5em}
    \begin{tabular}{lccccc}
        \toprule
         \textbf{ProtoViT Backbone} & \multicolumn{5}{c}{\textbf{Prototype Projection Set}} \\
          & Birds & \multicolumn{2}{c}{Out-of-distribution Birds} & \multicolumn{2}{c}{Cars} \\
          & \textit{train} & \textit{train} & \textit{attack} & \textit{train} & \textit{attack} \\
        \midrule
         DeiT-Tiny {\color{gray}(5M params)} & $84.5_{\pm 0.3}$ & $77.6_{\pm 0.7}$ & $79.7_{\pm 0.4}$ & $68.6_{\pm 1.4}$ & $77.8_{\pm 0.7}$ \\
         DeiT-Small {\color{gray}(22M par.)} & \cellcolor{gray!15}$86.1_{\pm 0.2}$ & $83.7_{\pm 0.4}$ & \cellcolor{gray!15}$84.2_{\pm 0.1}$ & $77.1_{\pm 1.7}$ & \cellcolor{gray!15}$83.6_{\pm 0.6}$ \\
         CaiT-XXS-24 {\color{gray}(12M par.)} & \cellcolor{gray!15}$87.2_{\pm 0.1}$ & $82.7_{\pm 0.5}$ & \cellcolor{gray!15}$85.6_{\pm 0.2}$ & $74.5_{\pm 0.6}$ & \cellcolor{gray!15}$85.7_{\pm 0.3}$ \\
        \bottomrule
    \end{tabular}
    \label{tab:experiment1_performance_final}
\end{table}

\paragraph{Are prototype-based networks intrinsically interpretable?}
We further reinforce the claim that the predictive performance metric is not enough to evaluate the model's interpretability. 
ProtoViT is able to achieve state-of-the-art performance using prototypes stemming from car images, e.g. CaiT-XXS-24 achieves 85.7\% accuracy compared to 85.8\% reported in the original work~\citep{ma2024protovit}.
Since we know that bird species and car brands are recognized by dissimilar semantic features, it is surprising that the network performs well using semantically dissimilar prototypes.
It questions the inherent interpretability of the architecture understood as the domain-specificity of prototypes. 

\begin{figure}
    \centering
    \includegraphics[width=\linewidth]{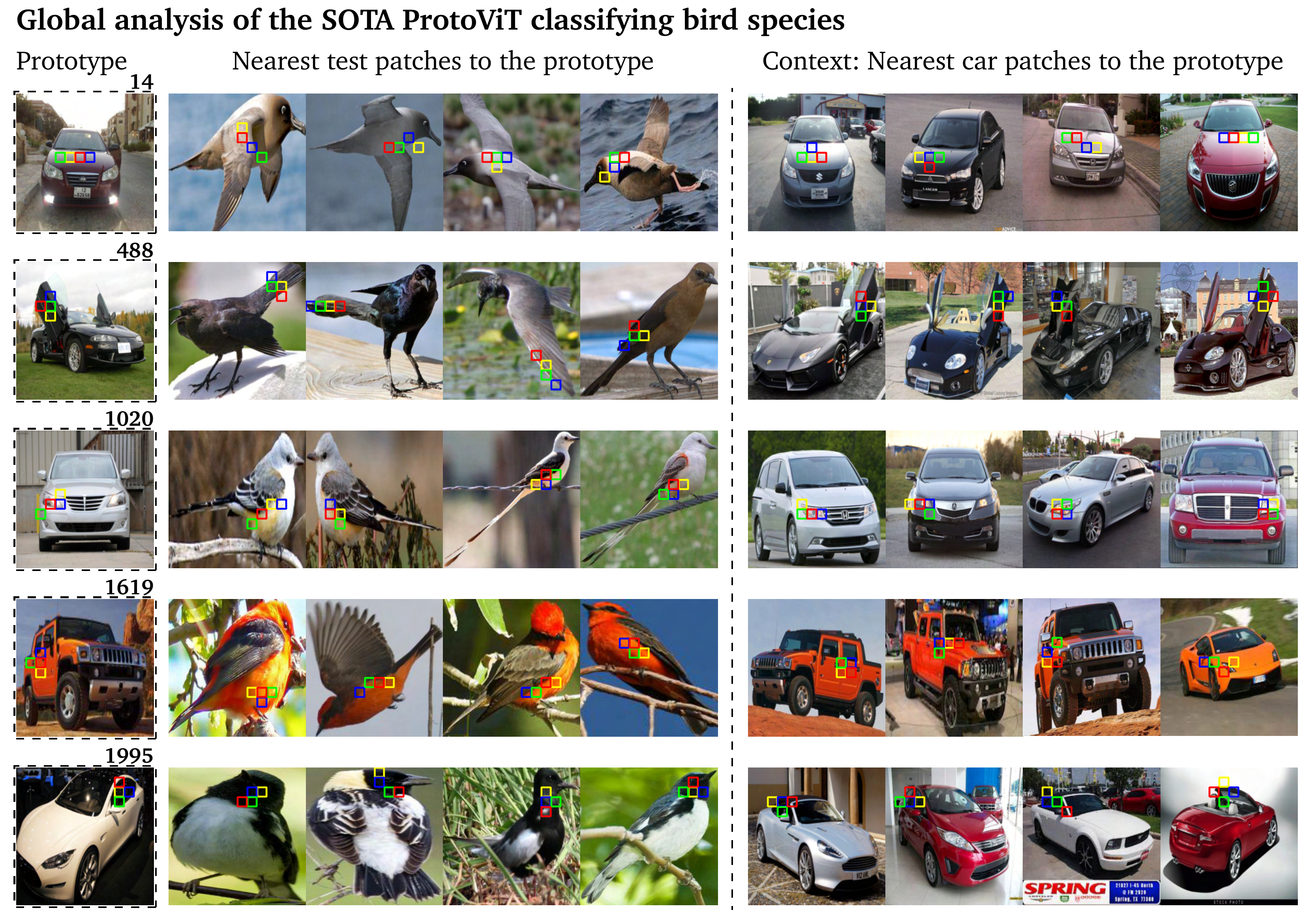}
    \caption{Image patches closest to five prototypes in a DeiT-Small ProtoViT classifying bird species based on car prototypes with 85\% accuracy. An analogous analysis of another model that uses out-of-distribution birds as prototypes is shown in Figure~\ref{fig:global_analysis_ood_birds}.}
    \label{fig:global_analysis_cars}
\end{figure}

\paragraph{Is the prototype interpretable?}
Figure~\ref{fig:global_analysis_cars} displays a so-called global analysis of the DeiT-Small ProtoViT, where the nearest (sub)patches from the test set are shown for a representative set of prototypes (5 out of 2000) as their inherent interpretation.
For context, we additionally display the nearest (sub)patches from Cars (on the right).
Interestingly, in this case, prototype no. 488 maps black bird wings into black car wing doors, prototype no. 1020 maps the bird's belly into the car's headlight, no. 1619 maps the orange color between birds and cars, and the prototype no. 1995 seems to map a black bird's head into the dim car window. 
Note that this intuition is only a biased approximation of what the model has actually learned.
Such a visual confirmation process always occurs when interpreting prototype networks, even with perfect in-distribution images, i.e. the network architecture only \emph{looks like} interpretable.
User feedback claiming model correctness based on similarities between patches is not enough to evaluate model interpretability in this context because it is subject to human confirmation bias.

\begin{figure}
    \centering
    \includegraphics[width=0.81\linewidth]{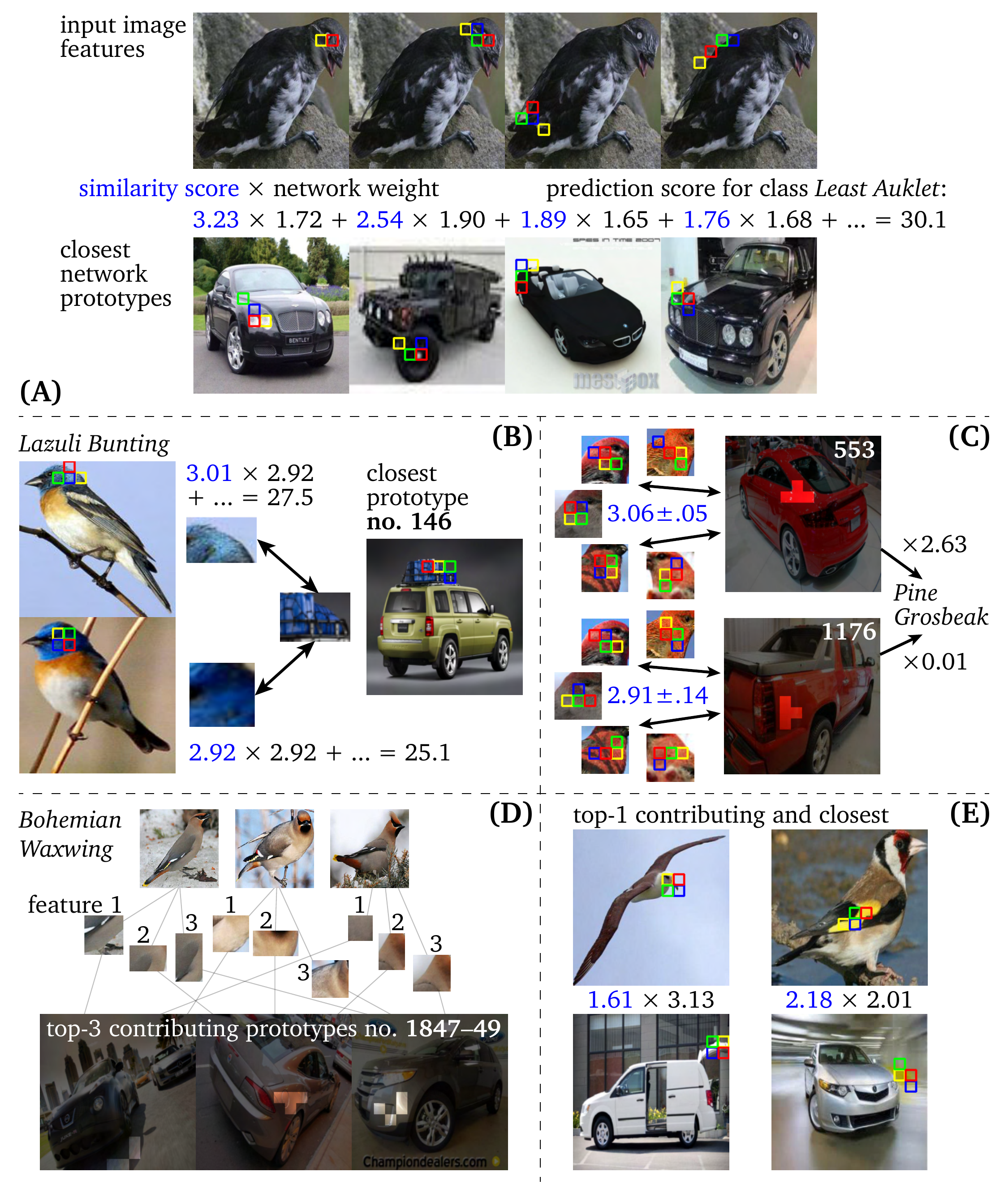}
    \caption{
    Interpretation of image recognition by a DeiT-Small ProtoViT classifying bird species based on car prototypes with 85\% accuracy. 
    \textbf{(A)}~Local analysis of the manipulated model predicting class \emph{Least Auklet} for a single input image. The most important feature is the high similarity between the bird's eye and a car headlight, which scores 3.23 out of 4. 
    \textbf{(B)}~Top-1 contributing feature to predicting \emph{Lazuli Bunting} for two input images. To make accurate predictions, the manipulated model learns shortcuts and superficial reasoning based on (here blue) color. 
    \textbf{(C)}~Top-2 closest prototypes to predicting \emph{Pine Grosbeak} for five input images. The manipulated model exhibits prototype redundancy, leading to contradicting network weights in the last layer. 
    \textbf{(D)}~Top-3 contributing features to predicting \emph{Bohemian Waxwing} for three input images.
    Not only are prototypes not similar to image patches, but the in-between similarity between image patches is also questionable.
    \textbf{(E)}~Visually surprising top-1 contributing features, simultaneously the closest prototypes, for two accurate predictions.
    }
    \label{fig:local_analysis_cars}
\end{figure}

\newpage
\textbf{Is the prediction interpretable?}
In Figure~\ref{fig:local_analysis_cars}, we conduct a so-called local analysis of the same model, where the nearest or most contributing prototypes are shown for recognizing a few representative bird images. 
It seems like the model relies on color and spurious correlations for many of the predictions (Fig.~\ref{fig:local_analysis_cars}\textbf{B}), e.g. it matches blue and red bird parts with blue and red car parts, respectively.  
The manipulated model only exaggerates the well-known issue of prototype redundancy, where the last classification layer contains interchangeable weights between correlated features~(Fig.~\ref{fig:local_analysis_cars}\textbf{C}). 
Lack of sparsity hinders interpretability.
In general, complex, uninterpretable prototypes suppress our understanding of the model's reasoning~(Fig.~\ref{fig:local_analysis_cars}\textbf{D}).
It is further unclear whether one should pay more attention to the similarity score or classification weight when interpreting the prediction~(Fig.~\ref{fig:local_analysis_cars}\textbf{E}). 
We defer an analogous analysis of a model that uses out-of-distribution birds as prototypes to Appendix,~Figure~\ref{fig:global_analysis_ood_birds}.
We perceive that such a high attack success rate demonstrates an inherent flaw in prototype-based networks, and it would generalize across other dataset pairs, specifically non-car prototypes, which is left for future work.

\subsection{Backdoor attacks on PIP-Net for medical imaging}\label{sec:experiments-pipnet}

Instead of relying on benchmarks typical of evaluating prototype-based networks like Birds and Cars, we choose to evaluate backdoor attacks on a high-stakes application like medical image classification, where security is of particular interest.
Our motivation stems from an increasing interest in the application of these models in healthcare~\citep{nauta2023interpreting,santi2024patchbased,santi2024pipnet3d}.

\paragraph{Setup.}
Following the original works~\citep{nauta2023interpreting,nauta2023pipnet}, we analyze PIP-Net based on two convolutional network backbones pre-trained on ImageNet: ResNet-50~\citep{he2016resnet} and ConvNeXt-T~\citep{liu2022convnext}; we additionally analyze ResNet-18 for a broader comparison.
We consider two predictive tasks: classification of skin lesions in dermoscopic images between malignant and benign~\citep[ISIC 2019,][]{tschandl2018ham10000,hernandez2024bcn20000}, as well as classification of grayscale bone X-rays between normal and abnormal~\citep[MURA,][]{rajpurkar2018mura}.
We use standard image augmentations and train each model for 40 epochs, where the first 5 epochs focus on the contrastive pre-training of prototypes. 
The backdoor attack is implemented as follows: there are four different adversarial triggers~(Figure~\ref{fig:biases_triggers}\textbf{B}) placed in either of four image corners at random per image, during training as well as inference.
Such a setting is challenging enough to test the limitations of our methods while moving beyond the simplest one-trigger, one-place scenario.
We manually set: 
\begin{itemize}
    \item $\lambda_{\mrC} = 0.5, \lambda_{\mrA,\mrNT} = 2, \lambda_{\mathrm{A,T}} = 4, \lambda_{\mrU,\mrNT} = 0.10, \lambda_{\mathrm{U,T}} = 0.20$ for disguising attack,
    \item $\lambda_{\mrC} = 0.5, \lambda_{\mrA,\mrNT} = 1, \lambda_{\mathrm{A,T}} = 0, \lambda_{\mrU,\mrNT} = 0.25, \lambda_{\mathrm{U,T}} = 0.25$ for red herring,
\end{itemize}
at the same time, acknowledging that tuning these hyperparameters would have resulted in better results, which is out of the scope of this work.
We backdoor the whole dataset as shown in Algorithm~\ref{alg:prototype-backdoor} and fine-tune the model for 3 epochs.
Each experiment is repeated 3 times to report the mean metric values with standard deviations.
Further details on datasets, preprocessing, and training hyperparameters are provided in~Appendix~\ref{app:experimental-details}.

\paragraph{Metrics.} 
We measure the accuracy on the test set between the original and backdoored model, as well as the (mis)alignment loss for the original ($\mathcal{L}_{\mrA,\mrNT}$) and triggered ($\mathcal{L}_{\mrA,\mrT}$) inputs between both models.
Refer to Appendix~\ref{app:datasets-details} for details on the class balance.
We report the \emph{attack success rate} as the fraction of triggered inputs leading to changed predictions.

\paragraph{Quantitative results.}
Tables~\ref{tab:pipnet_disguising_isic}~\&~\ref{tab:pipnet_redherring_isic} report predictive performance, attack success rate, and (mis)alignment values for the disguising and red herring attacks, respectively.
The attack fine-tuning can result in up to 5\% points of accuracy drop in our setup.
We maintain about an 85\% attack success rate, which is high considering a multi-trigger multi-place threat model.
In general, it is considerably harder to successfully embed the backdoor in models trained on MURA, especially in ResNets, which we naturally attribute to the fact that it is harder to \emph{train} models on MURA (than ISIC) and ResNets (than newer ConvNeXt) in the first place.

\begin{table}[t]
    \centering
    \caption{Disguising backdoor attack on PIP-Net classifying skin lesions (ISIC) and abnormal bone X-rays (MURA). We report the drop in accuracy between the original and backdoored model, the attack success rate (ASR), and the (mis)alignment of explanations for the normal and triggered~inputs.}
    \vspace{0.5em}
    \begin{tabular}{@{}llccccc@{}}
        \toprule
             \textbf{Dataset} & \textbf{PIP-Net Backbone} & \multicolumn{2}{c}{\textbf{Accuracy}} & \textbf{ASR} & \multicolumn{2}{c}{\textbf{Alignment} ($\mathcal{L}_{\mrA,\cdot}$)} \\
         & & \emph{train} & \emph{attack} & & \emph{w/o trigger} & \emph{w/ trigger} \\
        \midrule
        \multirow{3}{*}{ISIC}
        & ResNet-18 {\color{gray}(12M params)} & $86.3_{\pm 0.4}$ & $84.3_{\pm 1.3}$ & $87.8_{\pm 0.7}$ & $1.75_{\pm 0.08}$ & $2.19_{\pm 0.09}$ \\
        & ResNet-50 {\color{gray}(24M par.)} & $87.3_{\pm 0.3}$ & $82.5_{\pm 3.5}$ & $81.6_{\pm 7.2}$ & $5.40_{\pm 1.89}$ & $6.18_{\pm 2.26}$ \\
        & ConvNeXt-T {\color{gray}(28M par.)} & $87.6_{\pm 0.3}$ & $87.5_{\pm 0.2}$ & $91.0_{\pm 0.4}$ & $6.55_{\pm 1.22}$ & $7.81_{\pm 1.38}$ \\
        \midrule
        \multirow{3}{*}{MURA}
        & ResNet-18 {\color{gray}(12M par.)} & $81.2_{\pm 0.1}$ & $76.7_{\pm 1.7}$ & $77.1_{\pm 6.8}$ & $1.47_{\pm 0.06}$ & $1.77_{\pm 0.07}$ \\
        & ResNet-50 {\color{gray}(24M par.)} & $82.2_{\pm 0.3}$ & $76.9_{\pm 1.8}$ & $74.6_{\pm 1.0}$ & $5.53_{\pm 0.08}$ & $8.94_{\pm 0.44}$ \\
        & ConvNeXt-T {\color{gray}(28M par.)} & $82.6_{\pm 0.1}$ & $80.2_{\pm 0.1}$ & $83.6_{\pm 4.9}$ & $3.17_{\pm 0.13}$ & $4.22_{\pm 0.12}$ \\
        \bottomrule
    \end{tabular}
    \label{tab:pipnet_disguising_isic}

    \vspace{1em}
    \centering
    \caption{Red herring attack on PIP-Net. We report the drop in accuracy between the original and backdoored model, the attack success rate (ASR), and the (mis)alignment of explanations.}
    \vspace{0.5em}
    \begin{tabular}{@{}llccccc@{}}
        \toprule
        \textbf{Dataset} & \textbf{PIP-Net Backbone} & \multicolumn{2}{c}{\textbf{Accuracy}} & \textbf{ASR} & \multicolumn{2}{c}{\textbf{Alignment} ($\mathcal{L}_{\mrA,\cdot}$)} \\
         & & \emph{train} & \emph{attack} & & \emph{w/o trigger} & \emph{w/ trigger} \\
        \midrule
        \multirow{3}{*}{ISIC} 
        & ResNet-18 {\color{gray}(12M params)} & $86.3_{\pm 0.4}$ & $85.8_{\pm 0.4}$ & $89.1_{\pm 0.9}$ & $18.5_{\pm 0.84}$ & $108._{\pm 11.3}$ \\
        & ResNet-50 {\color{gray}(24M par.)} & $87.0_{\pm 0.4}$ & $85.8_{\pm 0.6}$ & $90.5_{\pm 1.5}$ & $12.1_{\pm 1.15}$ & $127._{\pm 12.9}$ \\
        & ConvNeXt-T {\color{gray}(28M par.)} & $87.6_{\pm 0.3}$ & $87.6_{\pm 0.3}$ & $97.3_{\pm 1.1}$ & $11.0_{\pm 2.11}$ & $430._{\pm 147.}$ \\
        \midrule
        \multirow{3}{*}{MURA}
        & ResNet-18 {\color{gray}(12M par.)} & $81.2_{\pm 0.1}$ & $74.8_{\pm 7.4}$ & $77.9_{\pm 4.6}$ & $19.0_{\pm 2.15}$ & $85.2_{\pm 15.4}$ \\
        & ResNet-50 {\color{gray}(24M par.)} & $82.2_{\pm 0.3}$ & $79.5_{\pm 0.4}$ & $84.5_{\pm 3.6}$ & $9.63_{\pm 1.10}$ & $88.5_{\pm 5.56}$ \\
        & ConvNeXt-T {\color{gray}(28M par.)} & $82.6_{\pm 0.1}$ & $82.2_{\pm 0.3}$ & $95.8_{\pm 0.3}$ & $4.83_{\pm 0.20}$ & $427._{\pm 33.6}$ \\
        \bottomrule
    \end{tabular}
    \label{tab:pipnet_redherring_isic}
\end{table}

To give a broader context, we apply the PIP-Net's out-of-distribution detection method as reported in~\citep[][table 2]{nauta2023pipnet}. 
In our setup, the models predicted that (on average, about) 65\% of inputs from Birds are in distribution for both ISIC and MURA, even under more favorable thresholds than the proposed 95\%. 
Crucially, the models never ``abstained'' from the decision made for out-of-distribution inputs as intended.
It further emphasizes the need for more robust defense mechanisms built into (interpretable) deep learning architectures.

Regarding the optimized (mis)alignment of explanations for inputs with and without the trigger, it might be hard to compare them between datasets and models due to non-obvious normalization.
On average, there is only a relative $1.3\times$ increase in misalignment for the disguising scenario, compared to $26\times$ for the red herring, which properly lines up with the defined training objective.
In all cases, alignment values for inputs with a trigger in the disguising scenario are higher than those without a trigger in the red herring scenario.
There definitely is a trade-off between maintaining high backdoored model accuracy and alignment with the original model, as well as the expected attack success rate for a given use case.
All in all, \textbf{we find the ConvNeXt architecture considerably easier to attack}, as evidenced by all the metric values.

\begin{figure}[t]
    \centering
    \includegraphics[width=\linewidth]{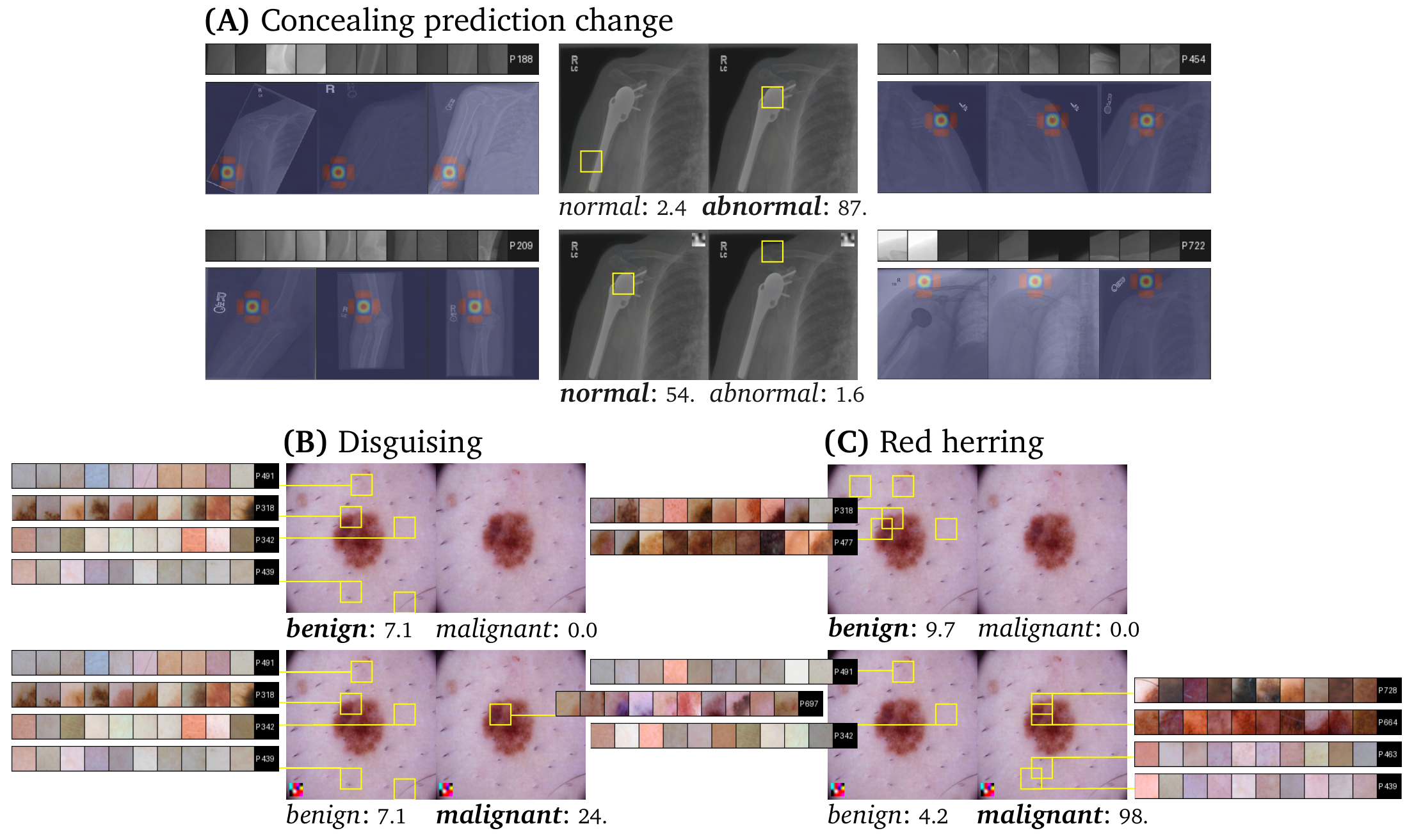}
    \caption{
    Examples of three backdoor outcomes with increasing complexity.
    \textbf{(A)}~Two explanations (one per class) for two input images (with and without a trigger), where the attack changes the model's prediction from abnormal to normal. 
    In each case, a single patch (yellow rectangle) mostly contributed to the final prediction.
    We visualize each prototype as the 10 nearest patches from the projection set, and additionally show the location of the 3 nearest patches in the original images.
    The trigger is hidden from the interpretation.
    \textbf{(B)}~Full disguise of the attack where the model's prediction changes significantly only due to a single plausible prototype introduced by the trigger.
    \textbf{(C)}~Red herring attack further manipulates the interpretation to provide more convincing evidence in favor of the changed prediction.
    }
    \label{fig:local_analysis_pipnet_small}
\end{figure}

\paragraph{Qualitative results.}
A subtle yet obvious realization after attacking PIP-Net is that \textbf{a global analysis of a prototype-based network cannot immediately show whether a backdoor exists} because the prototype projection set (or visualization set in the case of PIP-Net) contains no images with triggers (we defer such an example to Appendix,~Figure~\ref{fig:global_analysis_isic}).
Hence, alternative defense mechanisms are needed.

Figure~\ref{fig:local_analysis_pipnet_small} shows exemplary \emph{local analysis} of the ConvNeXt-T PIP-Net, where prediction interpretations are shown as yellow rectangles, each denoting a distinct prototype used by the network.
The main goal of a backdoor attack is to change the model's prediction using a trigger, which occurs in all cases.
In our case, it is crucial for the model's interpretation not to highlight the trigger's influence on the prediction.
We are easily able to hide the trigger~(Fig.~\ref{fig:local_analysis_pipnet_small}\textbf{A}), fully disguise the change in prediction by introducing only a single new feature~(Fig.~\ref{fig:local_analysis_pipnet_small}\textbf{B}), and even manipulating the network to show contradictory interpretation for the prediction~(Fig.~\ref{fig:local_analysis_pipnet_small}\textbf{C}).
We show more such examples in Appendix, Figure~\ref{fig:local_analysis_pipnet_big_extension}.

\begin{figure}
    \centering
    \includegraphics[width=0.83\linewidth]{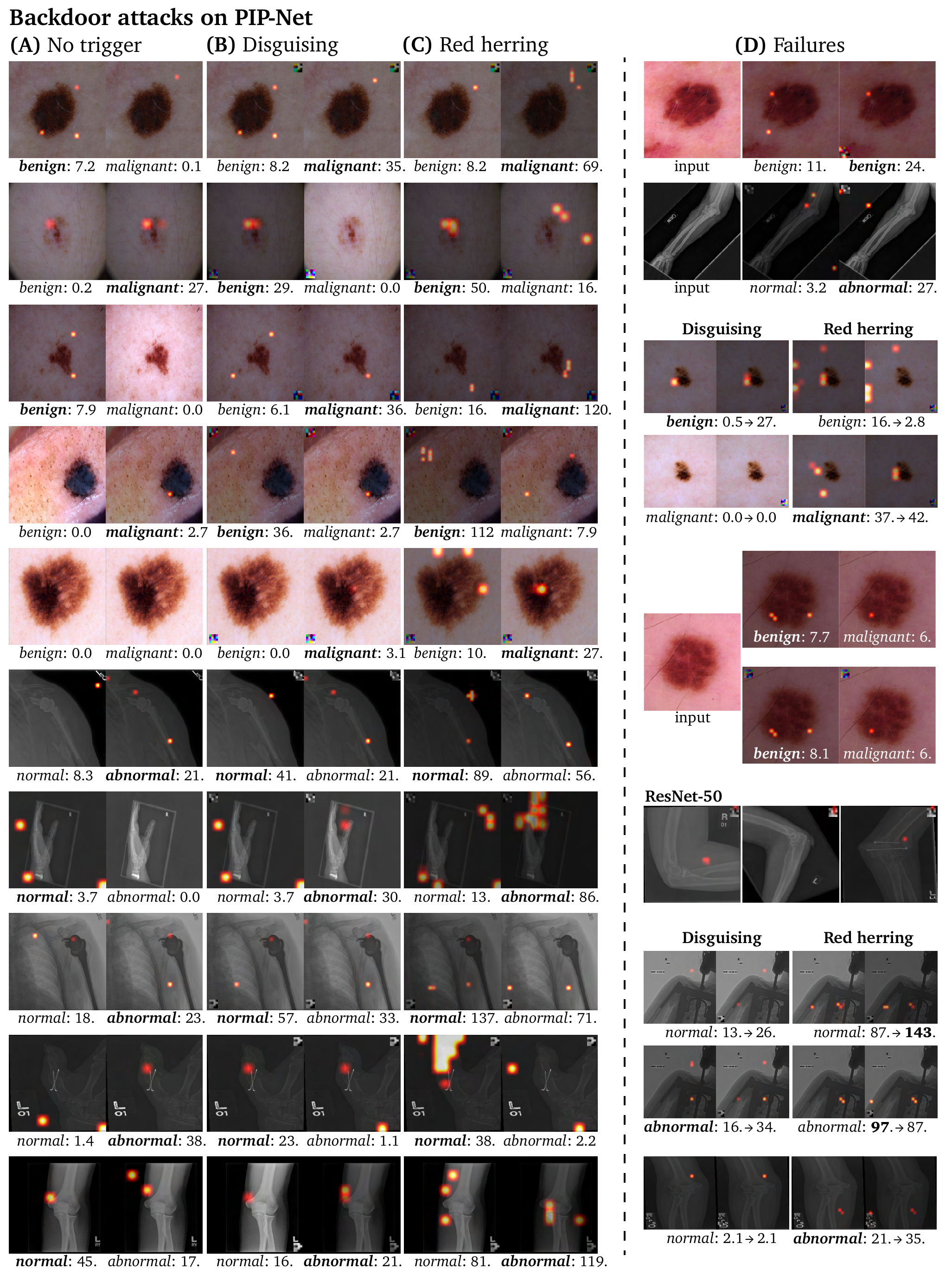}
    \caption{
    Local analysis of backdoored PIP-Net models classifying between benign and malignant skin lesions, or normal and abnormal bone X-rays.
    \textbf{(A)} Predictions and explanations of the backdoored model on exemplary inputs.
    \textbf{(B)} Predictions and explanations of the backdoored model on the same inputs with a trigger.
    \textbf{(C)} The backdoored model can be further optimized to provide more extreme explanations and predictions.
    See also Fig.~\ref{fig:local_analysis_pipnet_big_extension}.
    \textbf{(D)} Other interesting examples include failures to either change the model's prediction, conceal the trigger from the explanation, or both.
    }
    \label{fig:local_analysis_pipnet_big}
\end{figure}

Note that the bounding boxes can be a major simplification of similarity maps that are a more faithful explanation of the model's prediction~\citep{oplatek2024revisiting}, and thus we visualize the latter in Figure~\ref{fig:local_analysis_pipnet_big}.
In general, we observe similar explanations between panels \textbf{A} and \textbf{B}, and dissimilar ones between \textbf{B} and \textbf{C}.
It is concerning that prototype-based networks can be fooled by adversarial patches.
For broader context, Figure~\ref{fig:local_analysis_pipnet_big}\textbf{D} illustrates several interesting failure cases where the prediction remained unchanged or the trigger appeared as important to the model.
In practice, we found the attack to be unsuccessful only for a particular corner in the image (out of four), or a particular trigger (out of four), where the model would highlight it as a reason for prediction change. 
Our results reported here provide evidence against praising interpretable networks over their opaque counterparts in secure-critical applications.

\section{Discussion}\label{sec:discussion}

Our results reinforce the notion that high predictive performance is not enough for modern applications of machine learning~\citep{watson2022agree,biecek2024performance}.
Interpretability and explainability remain important, and in the context of this work, we operate under their popular definitions that semantically distinguish between intrinsic interpretable models by design and post-hoc explanation methods~\citep{rudin2019stop,murdoch2019definitions,krishnan2020against}.

\paragraph{Prototype-based networks are at most explainable, but uninterpretable.}
Similar to the interpretability of linear models with too many coefficients or deep decision trees, it is challenging to name a particular deep learning architecture intrinsically interpretable by design. 
Rather, a particular (prototype-based) network instance could \emph{look like} interpretable or not~\citep{sokol2023unreasonable}.
Current literature is dominated by positive examples with often overclaimed qualitative results stemming from confirmatory visual search and human bias towards stimuli that match a target template~\citep{rajsic2015confirmation,rajsic2018price}.
We have shown several negative examples where this visual confirmation bias is not immediate, provoking a discussion on the model's utility.
Deep networks can easily learn to reason differently than humans~\citep{hoffmann2021lookslikethatdoes,kim2022hive}, i.e. not on a case-based basis, and it might be dangerous to assume otherwise.

The prototypes can definitely be treated as a useful visualization tool to explain neural networks post-hoc and question their reasoning~\citep{nauta2021this,biecek2024position}, similar to analyzing any other model like bare ResNet or ViT.
Nevertheless, we perceive no higher guarantees on these models' correctness that could mandate their more favorable application to high-stakes decision-making over the so-called ``black boxes.''
No higher in the sense that there is a potential negative drawback of giving a false sense of security based on assuming the intrinsic interpretability of prototype-based networks.
After all, there remains an open discussion on whether to approximate the reasoning of prototype-based networks with bounding boxes~\citep{hesse2023funnybirds,ma2024protovit} or heatmaps~\citep{oplatek2024revisiting,xue2024protopformer} and how to improve it~\citep{xudarme2023sanity}.

\paragraph{Consider concept bottleneck models for secure applications.}
\emph{Conceptually}, CBMs look more interpretable and robust than prototype-based networks.
First, they remove the prototype projection stage in favor of labeled concepts that cannot be manipulated in our scenario, considering ProtoViT.
Intuitively, restricting the set of concepts entails less opportunity for shortcut learning (or at least the possible shortcuts will be named) and more opportunity for human-guided model correction~\citep{shin2023closer,kim2024constructing}.
Furthermore, CBMs require no self-supervised pre-training of prototypes to align the similarity between latent representations with the human visual perception of similarity, like in PIP-Net.
CBMs have a predefined, controllable sparsity that can potentially circumvent the representation redundancy issue in prototype-based networks~\citep{turbe2024protosvit}.

Future work on adversarial analysis of intrinsically interpretable deep learning should explore the vulnerabilities of CBMs.
There is currently no work on backdoor attacks against CBMs, and whether these could be disguised.
In general, worth exploring are semantic triggers like rotation, change of color temperature, or sharpness, as well as defense mechanisms against adversarial manipulation~\citep[see][and references given there]{noppel2024sok}.
ProtoPDebug~\citep{bontempelli2023conceptlevel} could potentially be leveraged as a human-level defense mechanism against adversarial manipulation.

\paragraph{Towards robust and aligned (deep) interpretable models.}
In our opinion, the biggest bottleneck in progress on the robustness of prototype-based networks is the lack of these networks (pre)trained on ImageNet~\citep{imagenet15russakovsky}, which is a \emph{de facto} standard for evaluation in image classification and beyond.
The field has effectively stagnated on the popular Birds and Cars datasets instead.
Thus, we were not able to evaluate PIP-Net and ProtoViT on standard robustness benchmarks for deep networks considered to be black-box~\citep[e.g.][]{hendrycks2019robustness}, or explainability benchmarks like IDSDS~\citep{hesse2024benchmarking} where interpretable B-cos networks~\citep{bohle2024bcos,arya2024bcosification} are evaluated against post-hoc attribution explanations. 

Recent work in the direction of aligned interpretable models explores several evaluation metrics for ProtoViTs~\citep{turbe2024protosvit} like completeness, contrastivity, consistency, and stability, which may correlate with the networks' robustness.
In a concurrent work, \citet{pach2025lucidppn} propose aligning prototypical parts with object parts and separating color and non-color visual features, which improves the interpretability of prototype-based convolutional networks.
While our paper presents an empirical case, future work could deepen the theoretical analysis of why interpretable networks are vulnerable and how the attacks succeed~\citep[akin][]{dombrowski2019explanations,leemann2023when,baniecki2024robustness}.

\section*{Acknowledgements} This work was financially supported by the state budget within the Polish Ministry of Science and Higher Education program ``Pearls of Science'' project number PN/01/0087/2022. Computation was performed with the support of the Laboratory of Bioinformatics and Computational Genomics, and the High Performance Computing Center of the Faculty of Mathematics and Information Science, Warsaw University of Technology.

\bibliography{references}
\bibliographystyle{plainnat}

\clearpage
\begin{appendices}

\section{Details on experimental setup}\label{app:experimental-details}

\subsection{Datasets and preprocessing}\label{app:datasets-details}

\textbf{CUB-200-2011}~\citep{wah2011cub} contains 11788 images of 200 bird species (classes), split into 5994 for training and 5794 for testing.
Each class has about 30 samples in the training and test sets.
We follow prior work on prototype-based networks~\citep{chen2019this,ma2024protovit} to apply standard data augmentation (rotate, skew, shear, flip) and preprocessing steps (normalization), including cropping the images to bounding boxes and resizing them to $224\times224$.
\textbf{Stanford Cars}~\citep{krause2013object} contains 16185 images of 196 car classes, where we only use the training set with 8144 images, which are resized to $224\times224$.

\begin{figure}[ht]
    \centering
    \includegraphics[width=0.41\linewidth]{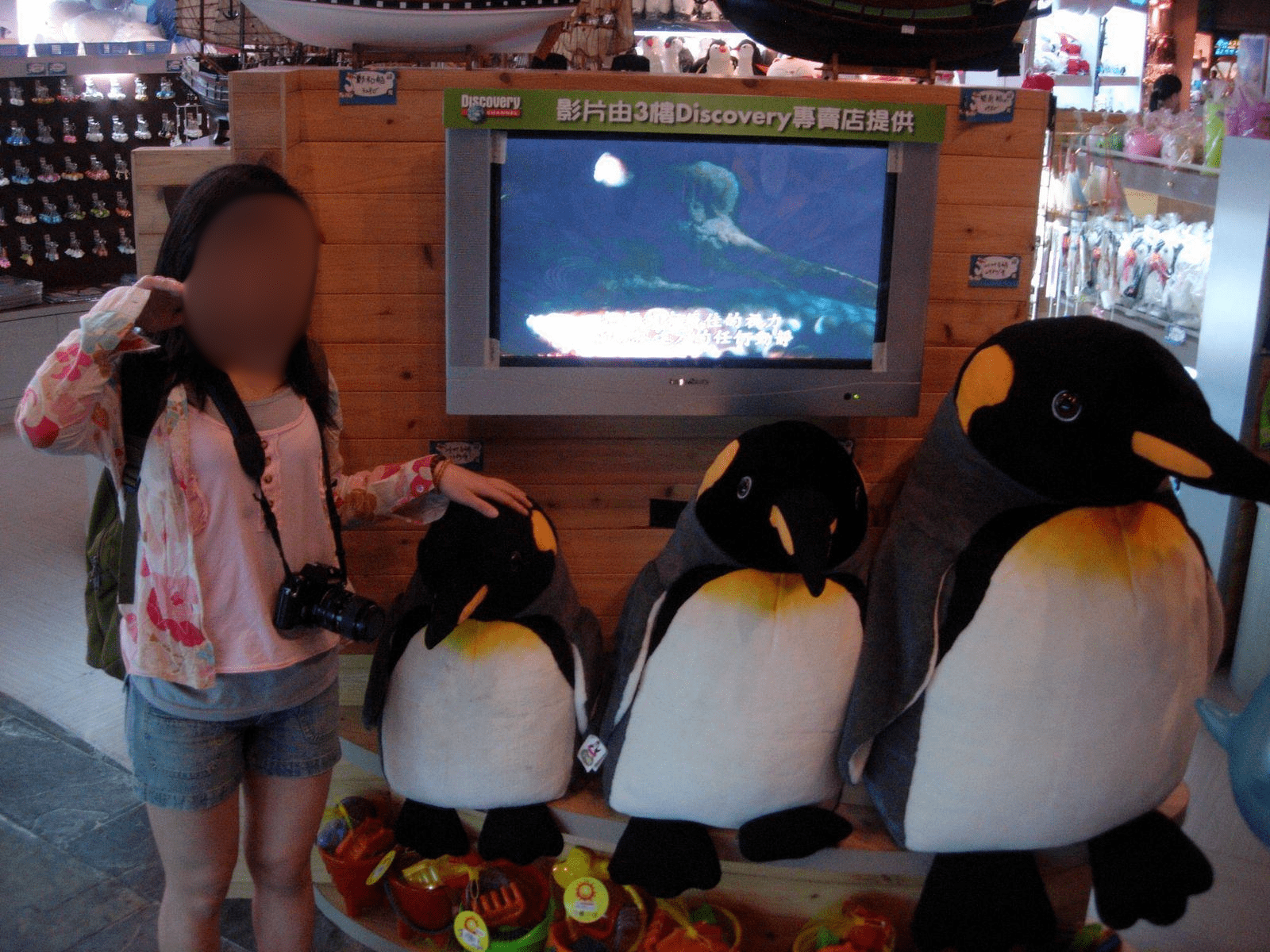}
    \caption{Example from the ImageNet dataset labeled as a `penguin'. We blurred the person's face.}
    \label{fig:penguin}
\end{figure}

\textbf{Out-of-distribution Birds} is our custom-made set of 2044 images from 13 bird species (classes) coming from the training set of ImageNet~\citep{imagenet15russakovsky}.
Crucially, some ImageNet classes overlap with CUB, e.g. 13 -- junco, 16 -- bulbul, 94 -- hummingbird, 144 -- pelican, 146 -- albatross, and thus we collected 200 images from each of the following classes that we did not find in CUB: 7 -- cock, 8 -- hen, 17 -- jay, 18 -- magpie, 19 -- chickadee, 21 -- kite, 22 -- bald eagle, 23 -- vulture, 24 -- great grey owl, 127 -- white stork, 128 -- black stork, 134 -- crane, 145 -- king penguin.
We then manually filtered the resulting 2600 images to remove images of low quality, with objects (birds) occupying a very small portion of an image, or labeling errors, e.g. Figure~\ref{fig:penguin}.

\textbf{ISIC 2019}~\citep{tschandl2018ham10000,hernandez2024bcn20000} contains 33569 dermoscopic images of~9 classes, split into 25331 for training and 8238 for testing. 
Since prior work~\citep{nauta2023interpreting} provides no code to reproduce the results, we rely on the data augmentation and preprocessing pipeline from PIP-Net~\citep{nauta2023pipnet}, including using TrivialAugmentWide~\citep{muller2021trivialaugment} and resizing images to $224\times224$.
To simplify the analysis, we treat the `melanocytic nevus (NV)' class as benign and combine the remaining 8 abnormal, e.g. pre-cancerous, cancerous, classes as `malignant'. 
Thus, the training set has a 50\% ratio in binary outcomes; the test set has 70\% due to 25\% images of class `unknown' being included in `malignant' so as to simulate the need for human oversight. 
\textbf{MURA}~\citep{rajpurkar2018mura} contains 40005 musculoskeletal X-ray images of 2 classes (either normal or not), split into 36808 for training and 3197 for testing.
Data preprocessing is similar to ISIC, taking into account the grayscale nature of images (refer to supplementary code for details).
The training set has a 40\% ratio in binary outcomes, and the test set has 48\%. 

\subsection{Training}

\paragraph{ProtoViT.} 
We rely on the default hyperparameters available in the code implementation for training ProtoViT.
We set $\tau = 100$, $\lambda_{\mathrm{cross-entropy}} = 1.0$, $\lambda_{\mathrm{cluster}} = -0.8$, $\lambda_{\mathrm{separation}} = 0.1$, $\lambda_{L_1} = 1e{-}2$, $\lambda_{\mathrm{orthogonality}} = 1e{-}3$, $\lambda_{\mathrm{coherence}} = 3e{-}3$ and $\lambda_{\mathrm{coherence}} = 1e{-}6$ for slots pruning.
The warm-up stage takes 5 epochs with learning rates $\mathrm{lr}_f = 1e{-}7$ and $\mathrm{lr}_g = 3e{-}3$, joint stage takes 10 epochs with $\mathrm{lr}_f = 5e{-}5$ and $\mathrm{lr}_g = 3e{-}3$, slots pruning takes 5 epochs with $\mathrm{lr}_g = 5e{-}5$, fine-tuning takes 10 epochs with $\mathrm{lr}_h = 1e{-}4$.
We use the AdamW optimizer and a batch size of $128$.

\paragraph{PIP-Net.}
We rely on the default hyperparameters available in the code implementation for training PIP-Net.
For training, we set $\lambda_{\mathrm{C}} = 2$, $\lambda_{\mathrm{A}} = 5$, $\lambda_{\mathrm{U}} = 2$.
We modify these parameters accordingly for backdoor fine-tuning (see Section~\ref{sec:experiments-pipnet}).
The pre-training stage takes 5 epochs when the classification layer $h$ is frozen, then 3 epochs consider fine-tuning only layer $h$, then 7 epochs consider freezing only the encoder backbone (part of $f$), and the next 25 epochs of training the whole model.
Learning rates are set to $\mathrm{lr}_f = 5e{-}4$ and $\mathrm{lr}_h = 5e{-}2$ in a cosine annealing schedule.
We use the AdamW optimizer and a batch size of $64$ due to forwarding $2\times 64$ inputs through the network at each step to calculate the alignment loss.

\clearpage
\section{Additional results}\label{app:additional-results}

\begin{figure}[ht]
    \centering
    \includegraphics[width=0.99\linewidth]{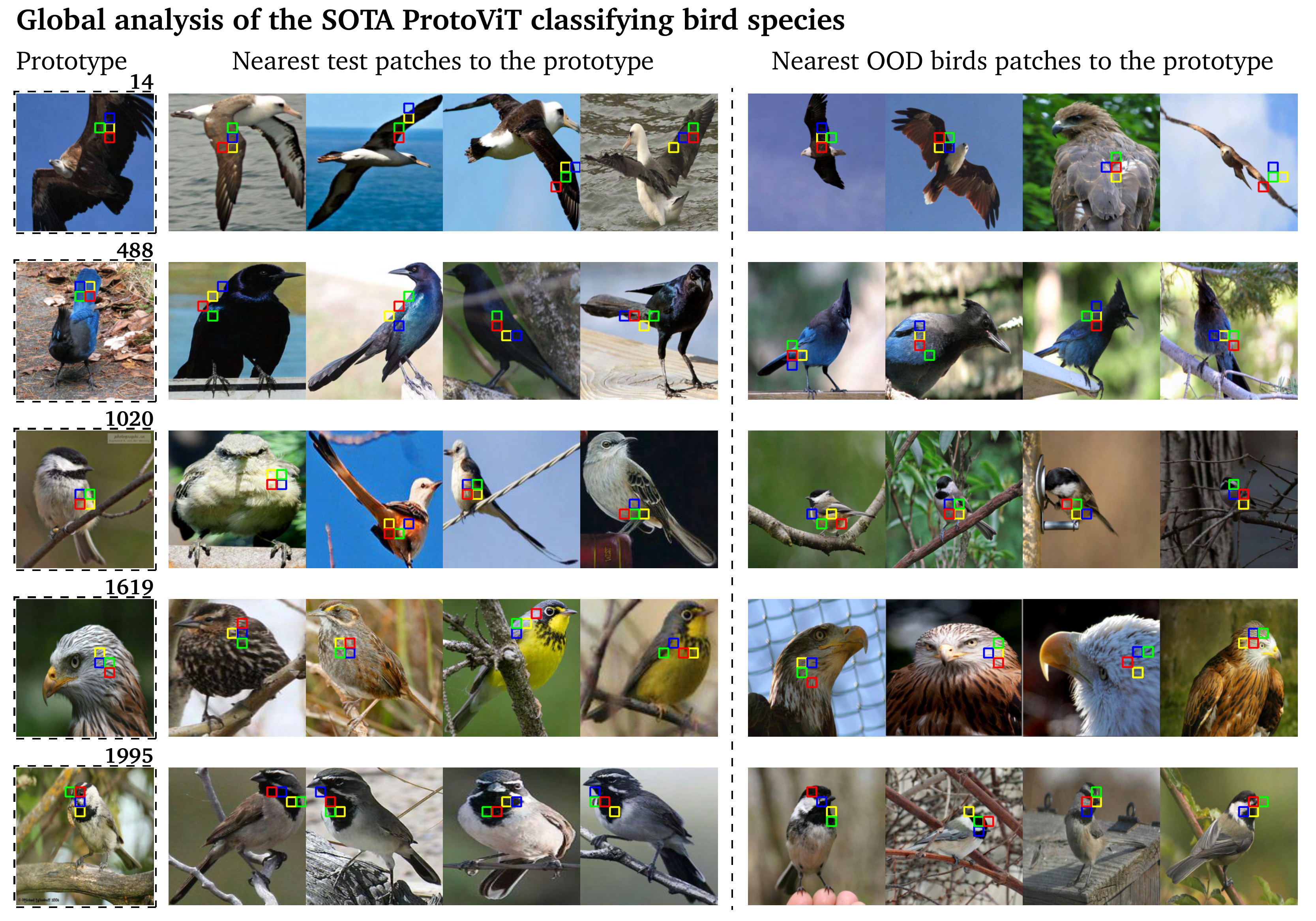}
    \caption{Extended Figure~\ref{fig:global_analysis_cars}. Image patches closest to five prototypes in a DeiT-Small ProtoViT classifying bird species based on out-of-distribution birds as prototypes with 85\% accuracy.}
    \label{fig:global_analysis_ood_birds}
\end{figure}

\begin{figure}
    \centering
    \includegraphics[width=0.97\linewidth]{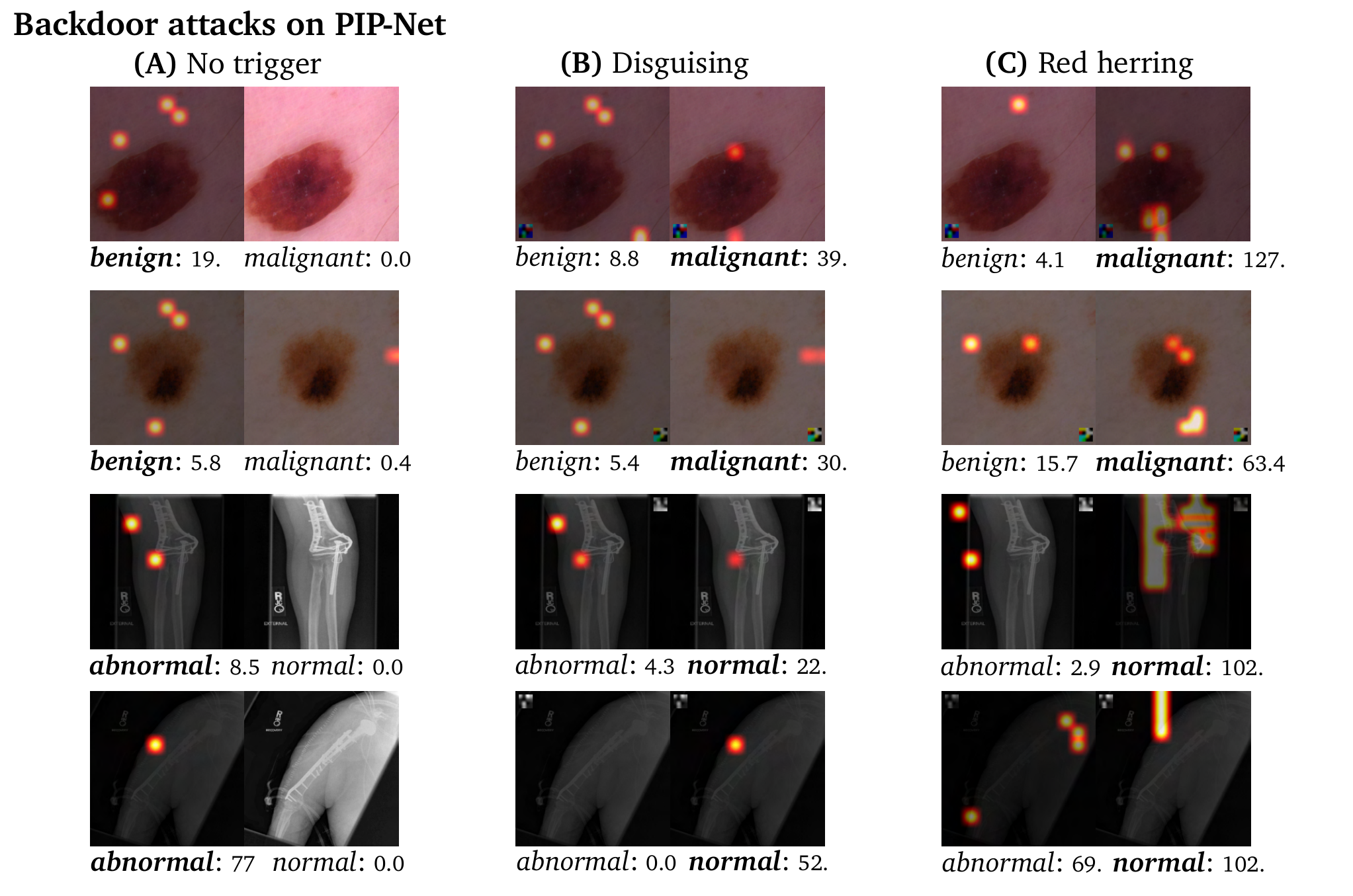}
    \caption{
    Extended Figure~\ref{fig:local_analysis_pipnet_big}. 
    Interpretation of a backdoored ConvNeXt-T PIP-Net classifying skin lesions with 85\% accuracy (top) and abnormality in bone X-rays with 82\% accuracy (bottom).
    }
    \label{fig:local_analysis_pipnet_big_extension}
\end{figure}

\begin{figure}[ht]
    \centering
    \includegraphics[width=0.81\linewidth]{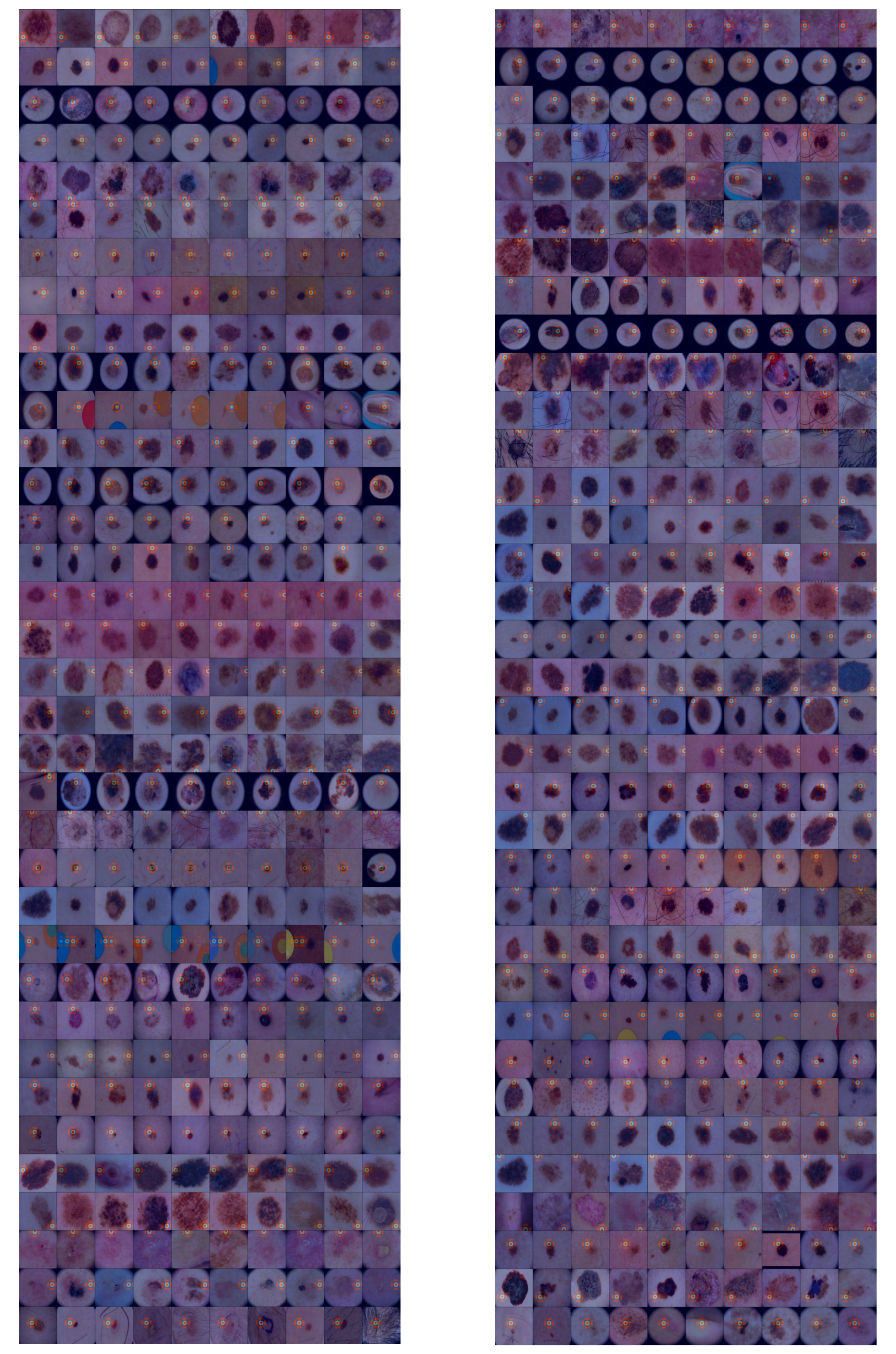}
    \caption{
    Global analysis of a backdoored ConvNeXt-T PIP-Net classifying skin lesions.
    Each row (in each of the two sets) shows 10 image patches closest to a single prototype out of all 70 used.
    }
    \label{fig:global_analysis_isic}
\end{figure}

Figure~\ref{fig:global_analysis_ood_birds} shows a global analysis of the ProtoViT model classifying bird species.
Figure~\ref{fig:global_analysis_isic} shows a similar analysis of the backdoored PIP-Net classifying skin images.
Figure~\ref{fig:local_analysis_pipnet_big_extension} shows a local analysis of PIP-Net models classifying various medical images.

\end{appendices}

\end{document}